\documentclass[final,3p,twocolumn,12pt,authoryear]{elsarticle}

\usepackage{amssymb}
\usepackage{amsmath}

 \usepackage{lineno}

\journal{Expert Systems with Applications}

\usepackage[]{acronym}
\usepackage{algorithm}%
\usepackage{algorithmicx}%
\usepackage{algpseudocode}%
\usepackage{listings}%
\usepackage[table]{xcolor}
\usepackage{graphicx}
\usepackage{adjustbox}
\usepackage{subcaption}
\usepackage{tikz}
\usepackage{longtable}
\usepackage{float}
\usepackage{collcell}
\usepackage{multirow}
\usepackage{booktabs}
\usepackage[figuresright]{rotating}
\usepackage{url}
\usepackage[pdfborder={0 0 0}]{hyperref}
\usepackage[capitalize]{cleveref}

\newcommand*{\MinNumber}{0}%
\newcommand*{\MaxNumber}{3}
\newcommand{\ApplyGradient}[1]{%
    \if\relax\detokenize{#1}\relax
    \else
      \pgfmathsetmacro{\raw}{100*(#1-\MinNumber)/(\MaxNumber-\MinNumber)}%
      \pgfmathtruncatemacro{\PercentColor}{max(0,min(100,round(\raw)))}%
      \edef\FullColor{darkgray!\PercentColor!white}%
      \expandafter\cellcolor\expandafter{\FullColor}{#1}%
    \fi

}
\newcolumntype{R}{>{\collectcell\ApplyGradient}c<{\endcollectcell}}

\newcommand{\plothistograms}[2]{%
    \centering
    \begin{minipage}[b]{0.24\textwidth}
        \centering
        \includegraphics[width=\textwidth]{output_#1-h-Otsu-#2.png}
    \end{minipage}
    \begin{minipage}[b]{0.24\textwidth}
        \centering
        \includegraphics[width=\textwidth]{output_#1-h-Kapur-#2.png}
    \end{minipage}
    \begin{minipage}[b]{0.24\textwidth}
        \centering
        \includegraphics[width=\textwidth]{output_#1-h-Kitler-#2.png}
    \end{minipage}
    \begin{minipage}[b]{0.24\textwidth}
        \centering
        \includegraphics[width=\textwidth]{output_#1-h-Kitler-met-overlapping-1.png}
    \end{minipage}

}

\newcommand{\plothistogramsn}[2]{%
    \centering
    \begin{minipage}[b]{0.24\textwidth}
        \centering
        \includegraphics[width=\textwidth]{output_hn_#1-h-Otsu-#2.png}
    \end{minipage}
    \begin{minipage}[b]{0.24\textwidth}
        \centering
        \includegraphics[width=\textwidth]{output_hn_#1-h-Kapur-#2.png}
    \end{minipage}
    \begin{minipage}[b]{0.24\textwidth}
        \centering
        \includegraphics[width=\textwidth]{output_hn_#1-h-Kitler-#2.png}
    \end{minipage}
    \begin{minipage}[b]{0.24\textwidth}
        \centering
        \includegraphics[width=\textwidth]{output_hn_#1-h-Kitler-met-overlapping-1.png}
    \end{minipage}

}

\newcommand{\plotthresholded}[2]{%
    \centering
    \begin{minipage}[b]{0.19\textwidth}
        \centering
        \includegraphics[width=\textwidth]{output_#1.png}
    \end{minipage}
    \begin{minipage}[b]{0.19\textwidth}
        \centering
        \includegraphics[width=\textwidth]{output_#1-Otsu-#2-seg.png}
    \end{minipage}
    \begin{minipage}[b]{0.19\textwidth}
        \centering
        \includegraphics[width=\textwidth]{output_#1-Kapur-#2-seg.png}
    \end{minipage}
    \begin{minipage}[b]{0.19\textwidth}
        \centering
        \includegraphics[width=\textwidth]{output_#1-Kitler-#2-seg.png}
    \end{minipage}
    \begin{minipage}[b]{0.19\textwidth}
        \centering
        \includegraphics[width=\textwidth]{output_#1-Kitler-met-overlapping-1-seg.png}
    \end{minipage}

}

\DeclareMathOperator*{\best}{best}

\acrodef{met}[MET]{Minimum Error Thresholding}
\acrodef{ssim}[SSIM]{structural similarity index measure}
\acrodef{psnr}[PSNR]{peak signal-to-noise ratio}
\acrodef{wcss}[WCSS]{within-cluster sum of squares}
\acrodef{chi}[CHI]{Calinskie-Harabasz Index}
\acrodef{dbi}[DBI]{Davies-Bouldin Index}
\acrodef{rs}[RS]{R-Sqaured}
\acrodef{rmsstd}[RMSSTD]{Root-mean squared standard deviation}
\acrodef{bsds500}[BSDS500]{Berkeley Segmentation Data Set}
\acrodef{ssd-uae}[SISSD]{satellite imagery semantic segmentation dataset}
\acrodef{nmi}[NMI]{Normalized Mutual Information}
\acrodef{brats2020}[BraTS20]{Brain Tumor Image Segmentation Benchmark 2020}
\acrodef{isic2016}[ISIC2016]{International Skin Imaging Collaboration 2016 dataset}

\begin{document}

\begin{frontmatter}



\title{A Dynamic Programming Framework for Discovering Count and Values of Multilevel Image Thresholding} 


\author[1]{Eslam Hegazy\corref{cor1}} 
\ead{eslam.sabry@guc.edu.eg}
\author[1]{Mohamed Gabr}
\ead{mohamed.kamel-gabr@guc.edu.eg}
\cortext[cor1]{Corresponding Author}
\affiliation[1]{
    department={Computer Science and Engineering Department},
    organization={German University in Cairo},
            city={New Cairo},
            postcode={11865},
            state={Cairo},
            country={Egypt}}

\begin{abstract}
Multilevel Image thresholding is an important preprocessing algorithm in computer vision applications nowadays. Since most common thresholding methods take the desired count of thresholds as input by the user, thresholding methods that automatically determines a suitable count of thresholds from the input image itself are advantageous. In this article, a novel thresholding method based on a dynamic programming algorithm and a modification of \acf{met} criterion is thoroughly presented. An empirical statistical study is performed to pinpoint why this proposed method is superior. Moreover, an extended comparison between this proposed method and other state-of-the-art methods is performed on a comprehensive set of natural, satellite and medical test images. The numerical results show that the proposed MET-DP method takes much less time than traditional dynamic programming thresholding methods when the number of thresholds is high. The proposed method can detect a suitable count of thresholds for most of tested images of different types. However, traditional methods that take the count of thresholds as input produce thresholded images of higher \ac{ssim} and \ac{psnr} values than MET-DP. Source code can be found on \url{https://w3id.org/met-dp/article1-code}.
\end{abstract}




\end{frontmatter}


\section{Introduction}
\label{sec:intro}

\begin{figure*}
    \centering
    \includegraphics[width=\linewidth]{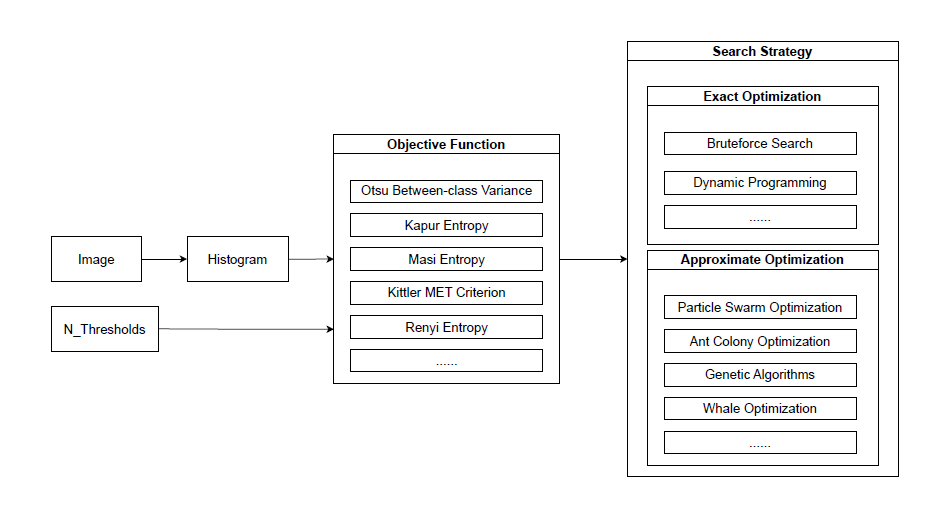}
    \caption{Framework for common thresholding methods. Inputs are a grayscale image (or the histogram of intensity levels of that image) and a count of thresholds. The two main components of a thresholding algorithm are the chosen objective function and search strategy}
    \label{fig:thresholding-framework}
\end{figure*}

Image thresholding is a common preprocessing technique for image processing and computer vision applications. Thresholding a grayscale image works by finding one (or more) intensity level(s) to separate image pixels into multiple groups. In case of applying one threshold, it is called bi-level thresholding. Similarly, applying two or more thresholds is called multilevel thresholding. Since the result of thresholding an image is a set of groups of pixels, image thresholding can be considered a special clustering problem.
State-of-the-art thresholding methods follow a common framework. This framework is illustrated in \cref{fig:thresholding-framework}. \Cref{fig:thresholding-framework} shows that a thresholding algorithm consists of an objective (evaluation) function and a search strategy. A thresholding algorithm following this framework takes as input (a histogram of) a grayscale image, and the desired count of thresholds to find. The thresholding method will apply the search method to find the set of thresholds that results in an optimal value of the evaluation function. This optimal value can be maximum or minimum depending on the chosen evaluation function.
Search methods can be classified into two categories: exact and approximate. Exact methods are guaranteed to reach an optimal value of the evaluation function. However, they usually suffer from exponential growth in runtime with increase in the desired count of thresholds. Approximate methods aim at improving (speeding up) the runtime by sacrificing optimality. Meaning that, they typically reach a suboptimal solution in a shorter runtime than exact methods that reach an optimal solution.

With the rise of metaheuristic optimization algorithms, approximate thresholding methods based on them are getting extensively studied recently. However, they are stochastic algorithms that involve randomized initialization and movements. This leads to these algorithms reaching different results on different runs. Also, they often suffer from getting stuck local minima or maxima. Also, the no-free-lunch theorem establishes that no optimization algorithm is dominant over other algorithms in all problem instances. Hence, if one thresholding method based on a metaheuristic algorithm will perform better than other methods based on other metaheuristic algorithms in some problems, it is to perform worse in other problems.
Another research direction is studying the thresholding methods based on exact search strategies. Although exact search strategies are usually slower than their approximate alternatives, their results are reliable since they do not involve randomness or stochasticity. Thresholding methods deploying dynamic programming date back to 1980 \citep{Kurita_Otsu_Abdelmalek_1992,otsu_1980_dp}. \citet{luessi2006,luessi2009} outlined a general framework for using dynamic programming with thresholding objective functions. This general framework showed efficient performance, specially for Otsu function.
\citet{lei_dp_masi} proposed a thresholding method that uses an adaptive variant of Masi entropy as an evaluation function and dynamic programming as a search strategy. This method achieved more optimal results than the competing methods in a comparable runtime. \citet{Hegazy_Gabr_B97_2025} conducted an extensive evaluation of that variant of Masi entropy over a large number of images and using different numbers of thresholds from 1 to 15. \citet{lei_dp_renyi} also presented a thresholding method based on dynamic programming and Renyi entropy. They did a transformation step on the input histogram to obtain its energy curve. This method showed results with superior thresholding quality as well as less time cost.

An earlier version of this work was presented in \citet{Hegazy_Gabr_B89_2025}. This article is an expansion of that work and contains material that has not been presented before. In particular, an extended theoretical analysis of the proposed method is presented in \cref{sec:methods}. Also, experimental testing has been performed on a bigger set of images and compared against more methods. Nonetheless, a guideline on use cases where our method is more or less useful than other methods is presented in \cref{sec:discussion}.

\section{Background}
\label{sec:background}

In this section, we provide the mathematical formulation of the multilevel image thresholding problem. This involves the mathematical equations for the objective functions and how their calculations are optimized.

\subsection{Thresholding Problem Formulation}
    \label{subsec:thr-prob-form}
Let the input grayscale image be represented by the function $f(x,y)$. Let $L$ be the number of allowed gray levels. In case of an 8-bit grayscale image, then $L= 2^8= 256$ intensity levels. Let the input image be of size $N \times M$. Therefore, the following inequality holds.

$0 \leq f(x,y) \leq L-1$ for $0 \leq x \leq N$ and $0 \leq y \leq M$

We define the histogram $h$ of this image $f(x,y)$ as a unary function that maps an intensity level to the number of occurrences of this intensity level in the input image. The number of occurrences of intensity level $i$ in the input image represented by the function $f$ is the size of the smallest set containing all pixels in the image having that intensity level $i$.

\begin{equation}
h(i)=| \{ (x,y) | f(x,y)=i \} |
\end{equation}

where $0 \leq i \leq L-1$.


We also define the normalized histogram $p$ as a function that maps an intensity level to the probability of a pixel in the image $f$ having that specific intensity level. $p(i)$ is computed by dividing $h(i)$ over the total number of pixels of the image. We denote the function $p(i)$ as $p_i$ for convenience.

\begin{equation}
    p_i=\frac{h(i)}{N \cdot M} \text{ for }0 \leq i \leq L-1
\end{equation}

For an arbitrary set of $r$ pixels $S=\{(x_1,y_1),(x_2,y_2),...(x_i,y_i),...,(x_r,y_r) \}$, we define the quantities $\mu_S, P_S, w_S, \sigma_S^2$ as follows.

    \begin{itemize}

        \item Number of pixels $P_S$
        \begin{equation}
            P_S = r 
        \end{equation}

        \item Weight $w_S$
        \begin{equation}
            w_S = \frac{P_S}{N \cdot M}
        \end{equation}

        \item Mean average intensity level $\mu_S$
        \begin{equation}
            \mu_{S}= \frac{\sum_{i=1}^{r} {\left ( f(x_i,y_i) \right )}}{P_S}
        \end{equation}

        \item Variance $\sigma_{S}^{2}$
        \begin{equation}
            \sigma_S^2 = \frac{1}{r} \cdot\sum_{i=1}^{r} {\left [f(x_i,y_i) -\mu_S \right ]^2}
        \end{equation}

    \end{itemize}

As these terms are to be used for calculating thresholding objective functions, we will rewrite these definitions for the case where $S = \{ (x,y) | a \leq f(x,y) \leq b\}$. This case is for a set of all pixels with intensity levels belonging to a closed interval $[a,b]$. Such a set constitutes a class/region of the histogram. We redefine these terms based on the normalized histogram function $p_i$ instead of the histogram function $h(i)$.

    \begin{itemize}

        \item Number of pixels $P_{[a,b]}$
        \begin{equation}
            P_{[a,b]} = \sum_{i=a}^{b}{h(i)}
        \end{equation}

        \item Weight $w_{[a,b]}$
        \begin{equation}
            \label{eq:weight-interval}
            w_{[a,b]} = \frac{P_{[a,b]}}{\sum_{i=0}^{L-1}{1}}
        \end{equation}

        \item Mean average intensity level $\mu_{[a,b]}$
        \begin{equation}
            \label{eq:mean-interval}
            \mu_{[a,b]}= \frac{\sum_{i=a}^{b} {\left ( i \cdot p_i \right )}}{w_{[a,b]}}
        \end{equation}

        \item Variance $\sigma_{[a,b]}^{2}$ and standard deviation $\sigma_{[a,b]}$
        \begin{equation}
            \label{eq:variance-interval}
            \sigma_{[a,b]}^2 = \frac{\sum_{i=a}^{b}{i \cdot i \cdot p_i}}{w_{[a,b]}} - \left (\mu_{[a,b]} \right )^2
        \end{equation}

        \begin{equation}
            \label{eq:std-interval}
            \sigma_{[a,b]} = \sqrt{\sigma_{[a,b]}^2}
        \end{equation}
    \end{itemize}

        Most common multi-level thresholding objective function are defined as a summation of values of a class cost function for each class. In other words, to evaluate $n$ thresholds that result in $n+1$ classes, a generalized formula for thresholding evaluation functions is the following.

        \begin{equation}
            J(T_1, ..., T_n) = \sum_{i=1}^{n+1}{Q(T_{i-1}, T_{i}-1)}
            \label{eqn:generic-eval-fun}
        \end{equation}

        We assume for simplicity that $T_0 = 0$ and $T_{n+1} = L$. We also assume that $0 = T_0 < T_1 < .... < T_n < L-1$. In \cref{eqn:generic-eval-fun}, $Q(T_{i-1}, T_{i}-1)$ is the cost function value of the i\textsuperscript{th} class (region). The i\textsuperscript{th} region contains all pixels with intensities in the closed interval $[T_{i-1}, T_{i}-1]$ (the same as $[T_{i-1}, T_i)$). The concrete definition of $Q(a,b)$ depends on the used thresholding objective function. We define such terms in \cref{eqn:otsu-q,eqn:kapur-q,eqn:kittler-q} for the objective functions of Otsu, Kapur and Kittler, respectively.


        \begin{equation}
            \label{eqn:otsu-q}
            Q(a,b)^{\text{Otsu}} = w_{[a, b]} \cdot \left ( \mu_{[a,b]} - \mu_{[0, L-1]} \right )^2
        \end{equation}


        \begin{equation}
            \label{eqn:kapur-q}
            Q(a,b)^{\text{Kapur}} = \log{w_{[a,b]}}- \frac{\sum_{j=a}^{b}{p_j * \log{p_j}}}{w_{[a, b]}}
        \end{equation}



        \begin{equation}
            \label{eqn:kittler-q}
            Q(a,b)^{\text{Kittler}} = w_{[a, b]} \cdot \left ( \log{\sigma_{[a, b]}} - \log{w_{[a, b]}} \right )
        \end{equation}

\subsection{Traditional Dynamic Programming for multi-level thresholding}
\label{subsec:trad-dp}

The common algorithm of dynamic programming onto multilevel thresholding is presented in \cref{alg:dp_n}. It takes as input the histogram of an image and the desired number of thresholds to be found. The algorithm works as follows. It builds a table $dp[i][j]$ that stores the minimum cumulative cost when the $i$-th threshold is placed at position $j$, considering all possible previous thresholds. The algorithm iteratively updates these values, tracking the best previous positions in $next[i][j]$. After filling the DP table, it selects the solution (set of thresholds) with the smallest total cost and reconstructs the optimal threshold positions by backtracking through $next$. This method efficiently finds globally optimal segmentation with a polynomial time complexity of $O(N \cdot L^2)$. It is designed for objective functions where the optimal solution corresponds to a minimum value, but it can also handle maximization objectives by reversing the comparison operator (using $>$ instead of $<$) and initializing $dp$ with $-\infty$ instead of $\infty$. The recurrence relation optimized by this algorithm is

        \begin{equation}
            \label{eqn:dp-n-recurrence}
            J(i, T_{j}) = \best_{T_k=i}^{T_j-1} \left ( J(i-1, T_k), Q(T_k, T_j) \right )
        \end{equation}

\begin{algorithm*}
    \caption{DP-N Algorithm}
    \label{alg:dp_n}

    \begin{algorithmic}[1]
    \State $dp[N\_Thresholds, L] \text{ initialized with } \mathtt{INF}$
    \State $next[N\_Thresholds, L] \text{ initialized with } -1$
    \For{$t = 0 : L-1$}
            \State $dp[0,t] \gets score(0, t)$
    \EndFor
    \For{$i = 1 : N\_Thresholds - 1$}
        \For{$T_j = i+1 : (L-1) - N\_Thresholds + i$}
            \For{$T_k = T_j - 1 \textbf{ down to } i $}
                \If{$dp[i-1][T_k] + score[T_k][T_j] < dp[i][T_j]$}
                    \State $dp[i][T_j] \gets dp[i-1][T_k] + score[T_k][T_j]$
                    \State $next[i][T_j] \gets T_k$
                \EndIf
            \EndFor
        \EndFor
    \EndFor

    \State $T_n \gets -1$
    \State $Best \gets \mathtt{INF}$
    \For{$i = 0 : L-1$}
        \If{$dp[N\_Thresholds-1][i] + score[i][L-1] < Best$}
            \State $Best \gets dp[N\_Thresholds-1][i] + score[i][L-1]$
            \State $T_n \gets i$
        \EndIf
    \EndFor
    \State $Thresholds \gets []$
    \State $t \gets T_n$
    \State $Thresholds.push(t)$
    \For {$i = N\_Thresholds - 2 \textbf{ down to } 1 $}
        \State $t \gets next[i][t]$
        \State $Thresholds.push(t)$

    \EndFor

    \end{algorithmic}
\end{algorithm*}

\section{Methods}
\label{sec:methods}

\subsection{Number of thresholds and thresholding quality}

It is a common practice when comparing the quality of different thresholding solutions, to use image similarity metrics between the original image and the thresholded image. \ac{ssim} and \ac{psnr} are of the most commonly used metrics for this purpose \citep{Amiriebrahimabadi_Rouhi_Mansouri_2024}. Therefore, increasing the number of thresholds allow more intensity levels in the thresholded image, making it more similar to the original image and hence, higher values for similarity metrics such as \ac{ssim} and \ac{psnr}. \ac{ssim} and \ac{psnr} will reach their maximum theoretical values when the original and thresholded images are identical. Thresholding results in an identical image to the original when there are enough thresholds to separate each existent intensity level in the image from all other existent intensity levels. For an image that contain all intensity levels in the interval $[0,L-1]$, $L-1$ thresholds are needed to separate the $L$ intensity levels. At the theoretical case of $0$ thresholds, all pixels of the output image have the same intensity level, values of \ac{ssim} and \ac{psnr} will be at their minimum over the input image possible thresholdings. 
In general, increasing the number of thresholds from $0$ to $L-1$ allows the thresholded image to be more similar to the input image and therefore, attaining higher values for similarity metrics.

Hence, for determining the appropriate number of thresholds for some input image, similarity metrics cannot be directly used to compare thresholding quality over different counts of thresholds. This is analogous to the problem of determining the number of clustering in general clustering \citep{McCullagh_Yang_2008,Xu_Qiao_Zhu_Zhang_Xue_Li_2016}. Increasing the number of clusters leads to each cluster being more compact. Hence, some metrics are biased into having more clusters/thresholds as it leads to a better separation/thresholding quality.
In clustering, there does not exist a universal method to determine the appropriate number of clusters for a given input dataset \citep{SAPUTRA2020}. However, multiple methods exist with each one having its own suitable use cases. These method include maximizing Dunn index, maximizing \ac{chi}, minimizing \ac{dbi}, or using the elbow method with \ac{rs} or \ac{rmsstd} or \ac{wcss}
\citep{Hassan_Tayfor_Hassan_Ahmed_Rashid_Abdalla_2024}.

\subsection{Experiment on the values of thresholding criteria functions changes with respect to number of thresholds}
\label{subsec:exp}

\begin{figure*}
    \centering
    \begin{subfigure}[]{0.48\linewidth}
        \includegraphics[width=\linewidth]{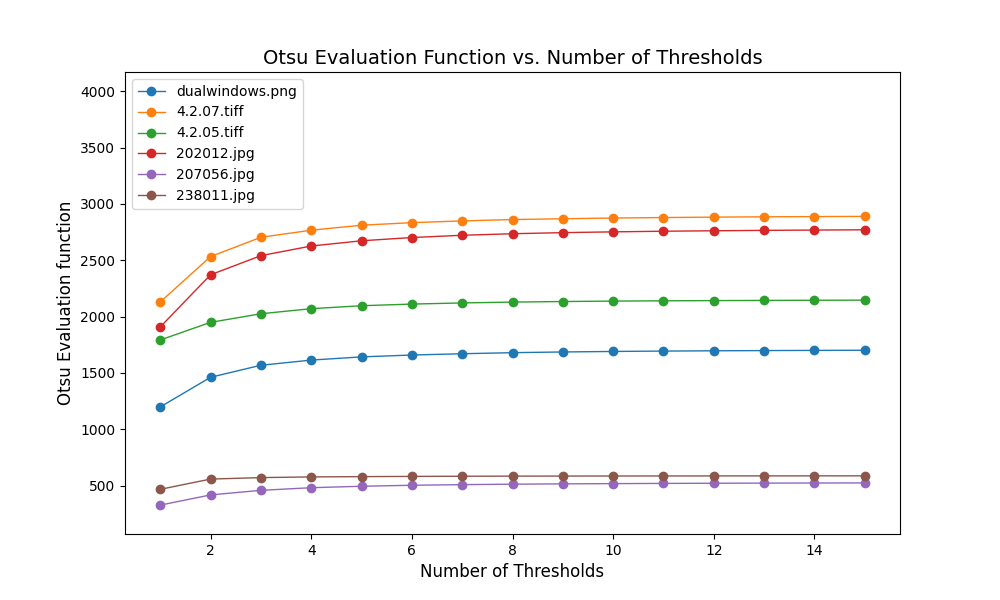}
        \caption{Otsu method (disjoint)}
        \label{fig:otsu-disj}
    \end{subfigure}
    \hfill
    \begin{subfigure}[]{0.48\linewidth}
        \includegraphics[width=\linewidth]{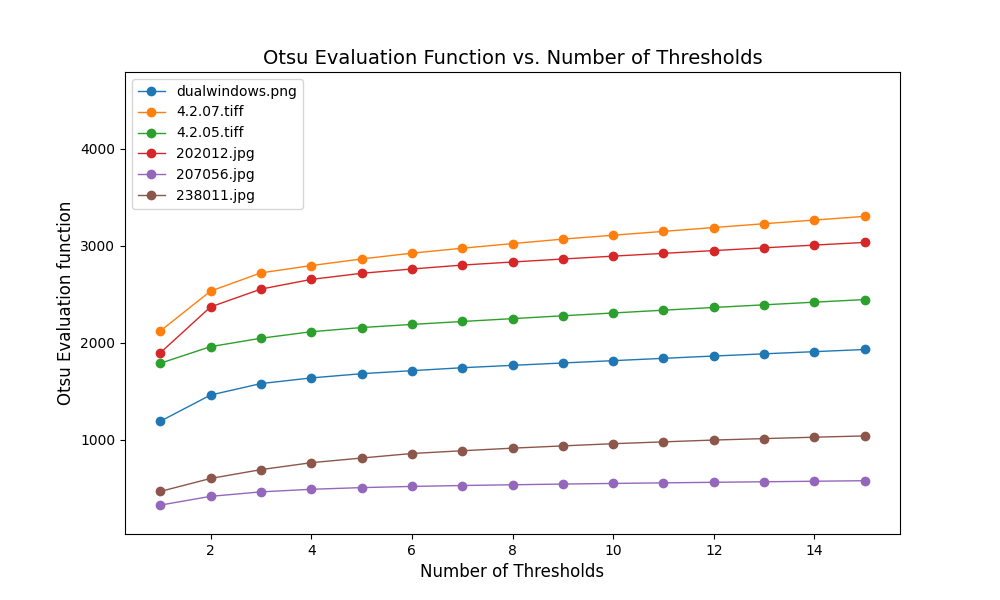}
        \caption{Otsu method (overlapping)}
        \label{fig:otsu-over}
    \end{subfigure}
    \begin{subfigure}[b]{0.48\linewidth}
        \includegraphics[width=\linewidth]{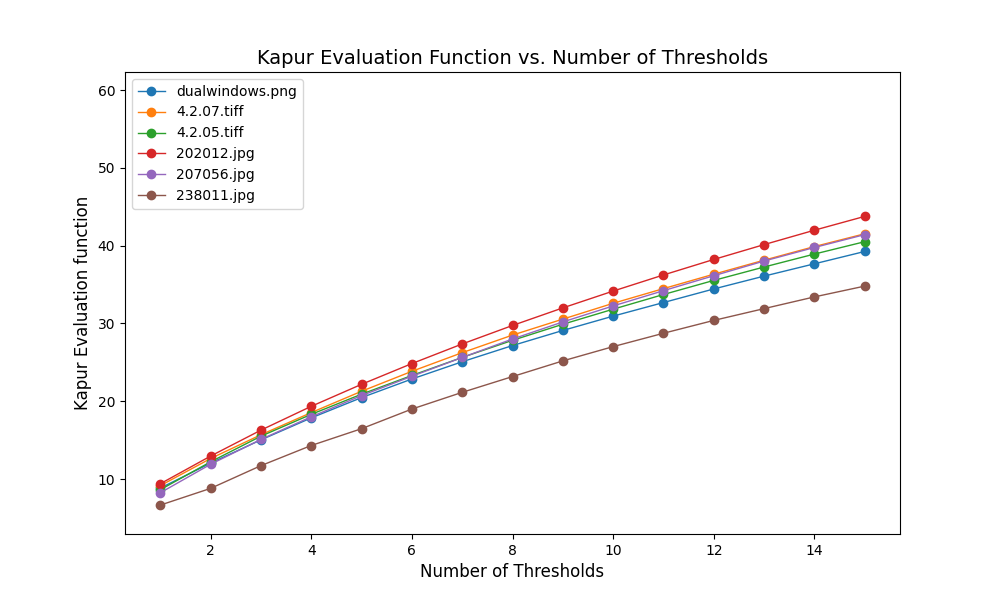}
        \caption{Kapur method (disjoint)}
        \label{fig:kapur-disj}
    \end{subfigure}
    \hfill
    \begin{subfigure}[b]{0.48\linewidth}
        \includegraphics[width=\linewidth]{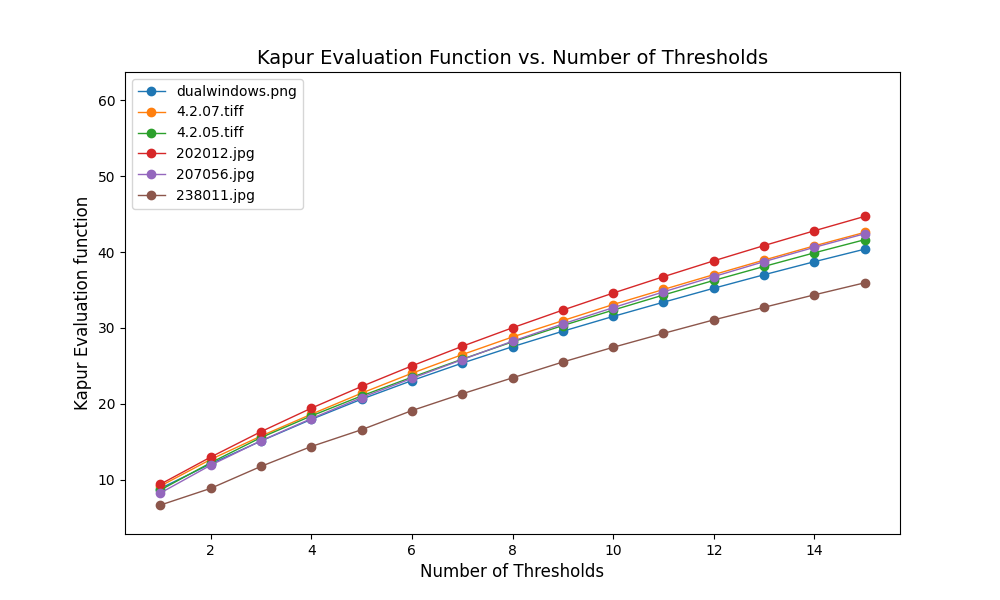}
        \caption{Kapur method (overlapping)}
        \label{fig:kapur-over}
    \end{subfigure}
     \begin{subfigure}[b]{0.48\linewidth}
        \includegraphics[width=\linewidth]{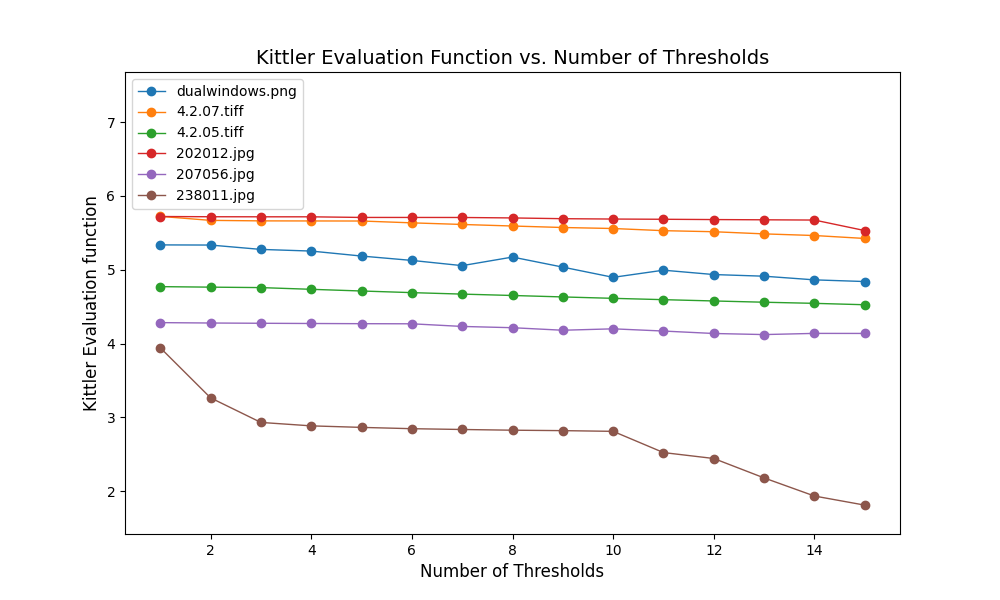}
        \caption{Kittler method (disjoint)}
        \label{fig:kittler-disj}
    \end{subfigure}
    \hfill
    \begin{subfigure}[b]{0.48\linewidth}
        \includegraphics[width=\linewidth]{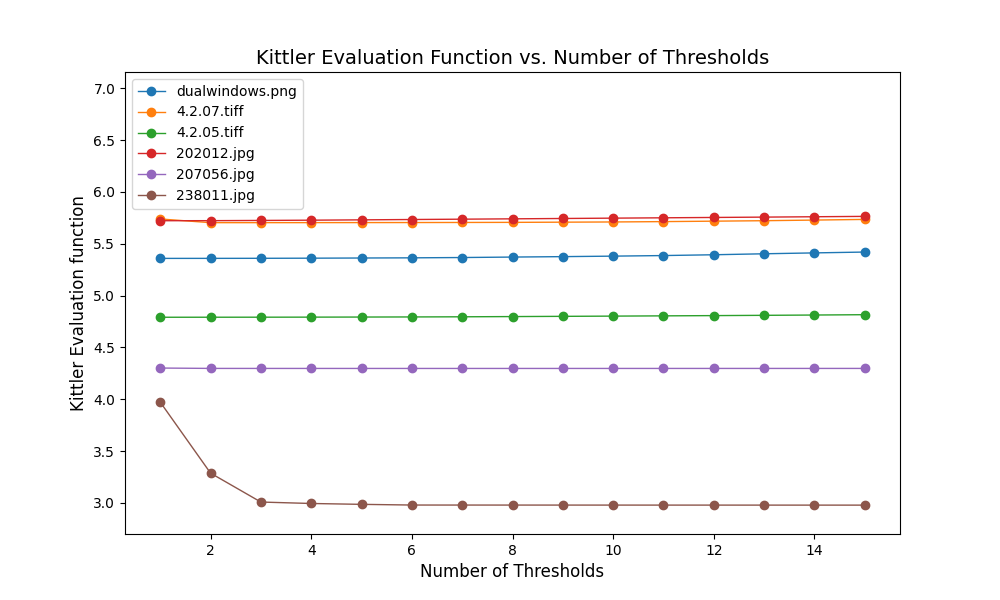}
        \caption{Kittler method (overlapping)}
        \label{fig:kittler-over}
    \end{subfigure}

    \caption{Optimal values of the Otsu, Kapur and Kittler fitness function across threshold numbers from 1 to 15, shown for five representative sample images. Subfigures in the left column show values for the standard (disjoint) formulation while subfigures in the right column show values for the overlapping formulation described in \cref{subsubsec:mod-met-criterion}.}
    \label{fig:exp1}
\end{figure*}

A quick experiment is performed over the Otsu, Kapur and Kittler evaluation functions. The traditional dynamic programming approach in \cref{subsec:trad-dp} is utilized. The experiment is performed over all counts of thresholds ranging from 1 to 15. Tested images are 500 images from the \ac{bsds500} \citep{bsds500}, 200 images from Weizmann segmentation evaluation database \citep{Weizmann} 
and 18 standard images (such as mandril and cameraman images).
\begin{table}
    \centering
    \begin{tabular}{|c|c|c|}
    \hline
         Image Name& Source & Notes \\
         \hline
    dualwindows & Weizmann &  \\
    eagle052607 & Weizmann & \\
    4.2.07 & Standard & Peppers\\
    4.2.05 & Standard & Jetplane\\
    202012 & \ac{bsds500} & Training set \\
    207056 & \ac{bsds500} & Training set\\
    238011 & \ac{bsds500} & Training set\\
    \hline
    \end{tabular}
    \caption{Details of source datasets of images used in \cref{fig:exp1}}
    \label{tab:exp-samp-img}
\end{table}
For each image, dynamic programming is used to find an optimal set of $n$ thresholds. After getting the best set of thresholds for every possible count from 1 to 15, the change of the value of the objective function between each two consecutive counts of thresholds is calculated.
Following this, we examine for each method separately the direction of change in the evaluation function. \Cref{tab:exp-change-disjoint} show the summary of the experiment results for each of Otsu, Kapur and Kittler functions respectively. For Otsu function, the change of Otsu function value from the optimal $m$ thresholds to the optimal $m+1$ thresholds (for $1 \leq m < 15$) is always positive.
\Cref{fig:exp1} show the actual evaluation function values over 5 sample images. Details about these five images are in \cref{tab:exp-samp-img}. \Cref{fig:otsu-disj} shows actual Otsu function values instead of changes. The point to deliver here is that Otsu values undergo a general trend of increase with increasing the number of thresholds. This is similar to the increase in \ac{ssim} or \ac{psnr} values. Hence, we cannot directly use Otsu function values to pick a suitable number of thresholds, as maximizing it will be biased towards the biggest number of thresholds possible and vice versa.
The same trend can be observed for Kapur function, as shown in \cref{tab:exp-change-disjoint,fig:kapur-disj}. However, examining Kittler values, we can make a different observation. In \cref{tab:exp-change-disjoint}, we see change amounts can be positive and can be negative. In \cref{fig:kittler-disj}, we see that the image, \textsc{dualimage.png} has its Kittler value at $n=7$ smaller than its Kittler value at $n=7$ while bigger than its value at $n=9$. Therefore, this suggests the possibility that $n=10$ which has the smallest Kittler value for this image is an appropriate number of thresholds. The point to deliver here is that Kittler criterion function does not undergo a total decrease in values in the same way Otsu and Kapur functions values increase with respect to the increase in number of thresholds.




\renewcommand{\arraystretch}{1}
\begin{table*}
\scriptsize
\centering
\begin{tabular}{|c|c|c|c|c|c|}
\hline
Method & Threshold Change & Min & Max & Avg & Std Dev \\
\hline
\multirow{14}{*}{Otsu} & 1 $\rightarrow$ 2 & 5.2786 & 1148.5165 & 384.3559 & 211.9557 \\
 & 2 $\rightarrow$ 3 & 3.7191 & 378.5974 & 132.8151 & 64.6779 \\
 & 3 $\rightarrow$ 4 & 1.6588 & 160.7974 & 62.6458 & 27.7096 \\
 & 4 $\rightarrow$ 5 & 1.0286 & 76.2916 & 34.2614 & 14.3778 \\
 & 5 $\rightarrow$ 6 & 0.8973 & 43.0337 & 20.8722 & 8.4567 \\
 & 6 $\rightarrow$ 7 & 0.1712 & 28.4637 & 13.6902 & 5.3761 \\
 & 7 $\rightarrow$ 8 & 0.1741 & 18.5787 & 9.3799 & 3.5663 \\
 & 8 $\rightarrow$ 9 & 0.2172 & 13.4733 & 6.7792 & 2.5929 \\
 & 9 $\rightarrow$ 10 & 0.3365 & 9.7387 & 5.0291 & 1.8641 \\
 & 10 $\rightarrow$ 11 & 0.1725 & 7.5457 & 3.8296 & 1.4045 \\
 & 11 $\rightarrow$ 12 & 0.0497 & 5.5521 & 2.9834 & 1.1000 \\
 & 12 $\rightarrow$ 13 & 0.0729 & 4.2655 & 2.3763 & 0.8712 \\
 & 13 $\rightarrow$ 14 & 0.0478 & 3.4207 & 1.9160 & 0.6929 \\
 & 14 $\rightarrow$ 15 & 0.0590 & 2.8156 & 1.5851 & 0.5653 \\
\hline
\multirow{14}{*}{Kapur} & 1 $\rightarrow$ 2 & 2.0599 & 4.3439 & 3.4705 & 0.2029 \\
 & 2 $\rightarrow$ 3 & 1.9136 & 3.7504 & 3.1493 & 0.1679 \\
 & 3 $\rightarrow$ 4 & 1.6304 & 3.4150 & 2.9086 & 0.1541 \\
 & 4 $\rightarrow$ 5 & 1.4061 & 3.0196 & 2.7187 & 0.1432 \\
 & 5 $\rightarrow$ 6 & 1.2910 & 2.8787 & 2.5607 & 0.1348 \\
 & 6 $\rightarrow$ 7 & 1.1365 & 2.7304 & 2.4229 & 0.1315 \\
 & 7 $\rightarrow$ 8 & 1.0286 & 2.5166 & 2.2987 & 0.1285 \\
 & 8 $\rightarrow$ 9 & 0.8884 & 2.4403 & 2.1882 & 0.1248 \\
 & 9 $\rightarrow$ 10 & 0.8268 & 2.3097 & 2.0884 & 0.1211 \\
 & 10 $\rightarrow$ 11 & 0.7179 & 2.1672 & 1.9989 & 0.1215 \\
 & 11 $\rightarrow$ 12 & 0.6118 & 2.1361 & 1.9158 & 0.1229 \\
 & 12 $\rightarrow$ 13 & 0.5011 & 2.0539 & 1.8392 & 0.1218 \\
 & 13 $\rightarrow$ 14 & 0.4970 & 1.9823 & 1.7698 & 0.1201 \\
 & 14 $\rightarrow$ 15 & 0.4187 & 1.8644 & 1.7032 & 0.1232 \\
\hline
\multirow{14}{*}{Kittler} & 1 $\rightarrow$ 2 & -1.2068 & 0.0006 & -0.0762 & 0.1226 \\
 & 2 $\rightarrow$ 3 & -0.9303 & 0.0254 & -0.0295 & 0.0549 \\
 & 3 $\rightarrow$ 4 & -1.1401 & 0.3253 & -0.0240 & 0.0733 \\
 & 4 $\rightarrow$ 5 & -1.6827 & 0.9506 & -0.0279 & 0.0944 \\
 & 5 $\rightarrow$ 6 & -2.2115 & 0.0845 & -0.0375 & 0.1379 \\
 & 6 $\rightarrow$ 7 & -1.9431 & 1.0725 & -0.0444 & 0.1468 \\
 & 7 $\rightarrow$ 8 & -1.7737 & 0.5078 & -0.0420 & 0.1429 \\
 & 8 $\rightarrow$ 9 & -2.2718 & 0.8627 & -0.0388 & 0.1547 \\
 & 9 $\rightarrow$ 10 & -2.1725 & 1.8627 & -0.0470 & 0.1736 \\
 & 10 $\rightarrow$ 11 & -2.8488 & 1.4056 & -0.0462 & 0.1749 \\
 & 11 $\rightarrow$ 12 & -1.4835 & 0.9006 & -0.0377 & 0.1359 \\
 & 12 $\rightarrow$ 13 & -2.0385 & 0.9096 & -0.0422 & 0.1470 \\
 & 13 $\rightarrow$ 14 & -2.9664 & 0.2921 & -0.0461 & 0.1641 \\
 & 14 $\rightarrow$ 15 & -2.3908 & 0.3830 & -0.0527 & 0.1625 \\
\hline
\end{tabular}
\caption{Statistical summary of the change in the optimal values of the fitness functions between consecutive counts of thresholds, $n$ and $n+1$ for $1 \leq n \leq 14$ over 718 images. Fitness functions here are calculated with a threshold value being included only in the next class. The reported values include the minimum, maximum, mean, and standard deviation for the Otsu, Kapur, and Kittler fitness functions.}
\label{tab:exp-change-disjoint}
\end{table*}
\renewcommand{\arraystretch}{1}

\renewcommand{\arraystretch}{1}
\begin{table*}
\centering
\scriptsize
\begin{tabular}{|c|c|c|c|c|c|}
\hline
Method & Threshold Change & Min & Max & Avg & Std Dev \\
\hline
\multirow{14}{*}{Otsu} & 1 $\rightarrow$ 2 & 15.8167 & 4529.6192 & 443.0393 & 362.3388 \\
 & 2 $\rightarrow$ 3 & 6.3925 & 1558.8428 & 203.5559 & 177.4231 \\
 & 3 $\rightarrow$ 4 & 4.3976 & 1501.3771 & 127.2391 & 114.0002 \\
 & 4 $\rightarrow$ 5 & 3.2910 & 760.5992 & 94.5782 & 79.5802 \\
 & 5 $\rightarrow$ 6 & 2.7601 & 659.1057 & 76.3575 & 63.7561 \\
 & 6 $\rightarrow$ 7 & 2.5370 & 470.9831 & 65.4942 & 52.3742 \\
 & 7 $\rightarrow$ 8 & 2.0241 & 393.5831 & 57.9681 & 45.5849 \\
 & 8 $\rightarrow$ 9 & 1.9919 & 378.3788 & 52.4546 & 41.0452 \\
 & 9 $\rightarrow$ 10 & 1.9504 & 257.9172 & 48.5587 & 37.0283 \\
 & 10 $\rightarrow$ 11 & 1.8412 & 242.6201 & 45.0504 & 33.9645 \\
 & 11 $\rightarrow$ 12 & 1.5384 & 232.6808 & 42.2628 & 31.4961 \\
 & 12 $\rightarrow$ 13 & 1.5318 & 213.8465 & 39.9539 & 29.4695 \\
 & 13 $\rightarrow$ 14 & 1.4542 & 186.6399 & 38.0020 & 27.2967 \\
 & 14 $\rightarrow$ 15 & 1.4080 & 179.5201 & 36.4857 & 26.0071 \\
\hline
\multirow{14}{*}{Kapur} & 1 $\rightarrow$ 2 & 2.0937 & 4.3200 & 3.4857 & 0.1992 \\
 & 2 $\rightarrow$ 3 & 1.9950 & 3.7614 & 3.1726 & 0.1653 \\
 & 3 $\rightarrow$ 4 & 1.7550 & 3.4294 & 2.9405 & 0.1491 \\
 & 4 $\rightarrow$ 5 & 1.5501 & 3.0353 & 2.7578 & 0.1378 \\
 & 5 $\rightarrow$ 6 & 1.4633 & 2.8904 & 2.6080 & 0.1286 \\
 & 6 $\rightarrow$ 7 & 1.3378 & 2.7789 & 2.4797 & 0.1240 \\
 & 7 $\rightarrow$ 8 & 1.2459 & 2.5685 & 2.3639 & 0.1199 \\
 & 8 $\rightarrow$ 9 & 1.1452 & 2.4828 & 2.2618 & 0.1153 \\
 & 9 $\rightarrow$ 10 & 1.0852 & 2.3729 & 2.1708 & 0.1125 \\
 & 10 $\rightarrow$ 11 & 1.0295 & 2.2451 & 2.0892 & 0.1088 \\
 & 11 $\rightarrow$ 12 & 0.9630 & 2.1730 & 2.0137 & 0.1087 \\
 & 12 $\rightarrow$ 13 & 0.8861 & 2.0952 & 1.9452 & 0.1076 \\
 & 13 $\rightarrow$ 14 & 0.8801 & 2.0848 & 1.8854 & 0.1036 \\
 & 14 $\rightarrow$ 15 & 0.7844 & 1.9754 & 1.8267 & 0.1048 \\
\hline
\multirow{14}{*}{Kittler} & 1 $\rightarrow$ 2 & -1.0991 & 0.0055 & -0.0612 & 0.1171 \\
 & 2 $\rightarrow$ 3 & -0.2789 & 0.0070 & -0.0112 & 0.0267 \\
 & 3 $\rightarrow$ 4 & -0.1238 & 0.0081 & -0.0024 & 0.0109 \\
 & 4 $\rightarrow$ 5 & -0.0567 & 0.0116 & -0.0001 & 0.0053 \\
 & 5 $\rightarrow$ 6 & -0.0630 & 0.0115 & 0.0009 & 0.0037 \\
 & 6 $\rightarrow$ 7 & -0.0537 & 0.0125 & 0.0014 & 0.0030 \\
 & 7 $\rightarrow$ 8 & -0.0195 & 0.0136 & 0.0018 & 0.0025 \\
 & 8 $\rightarrow$ 9 & -0.0003 & 0.0137 & 0.0021 & 0.0025 \\
 & 9 $\rightarrow$ 10 & -0.0004 & 0.0138 & 0.0024 & 0.0027 \\
 & 10 $\rightarrow$ 11 & -0.0035 & 0.0154 & 0.0027 & 0.0030 \\
 & 11 $\rightarrow$ 12 & -0.0017 & 0.0153 & 0.0029 & 0.0031 \\
 & 12 $\rightarrow$ 13 & -0.0001 & 0.0163 & 0.0032 & 0.0033 \\
 & 13 $\rightarrow$ 14 & -0.0001 & 0.0179 & 0.0033 & 0.0034 \\
 & 14 $\rightarrow$ 15 & -0.0001 & 0.0181 & 0.0035 & 0.0035 \\
\hline
\end{tabular}
\caption{Statistical summary of the change in the optimal values of the fitness functions between consecutive counts of thresholds, $n$ and $n+1$ for $1 \leq n \leq 14$ over 718 images. Fitness functions here are calculated with a threshold value being included in the previous and next class. The reported values include the minimum, maximum, mean, and standard deviation for the Otsu, Kapur, and Kittler fitness functions.}
\label{tab:exp-change-overlapping}
\end{table*}
\renewcommand{\arraystretch}{1}

%
%

\subsection{Modified \ac{met} based dynamic programming Algorithm}

    The proposed method consists of a dynamic programming algorithm variant, presented in \cref{alg:dp} and a modified version of \ac{met} criterion. We refer to this method as MET-DP method. For other methods that use the traditional dynamic programming variant explained in \cref{subsec:trad-dp} we refer to them as DP\_N\_Otsu or DP\_N\_Kapur or DP\_N\_Kittler, depending on the choice of objective function.

\subsubsection{Dynamic Programming Variant}

        \Cref{alg:dp} implements a dynamic programming approach to determine the optimal histogram thresholds by minimizing a given cost function. It maintains a 2D array $mem[i][j]$, where each entry represents the minimum achievable cost for treating the interval $[i, j]$ either as a single class or as multiple classes divided at some split point $k$. A companion array $next[i]$ stores the position of the optimal split for each starting index. The algorithm iterates backward through the histogram, updating $mem[i][j]$ whenever dividing the interval results in a lower total cost. Once all values are computed, the sequence of optimal thresholds is reconstructed by following the split positions recorded in $next$.

        The recurrence relation optimized by this dynamic programming algorithm is given by \cref{eqn:dp-recurrence}, where $Q(a,b)$ is the cost function for the region of intensity levels $[a,b]$ and $J$ is the total cost function. In general, $J$ can be a maximization or minimization function. Hence, $best$ mentioned in the equation is a shorthand for $\min$ or $\max$, depending on the nature of the concrete objective function optimized by the algorithm. In \cref{alg:dp}, we are assuming the objective function is to be minimized. Moreover, values of $Q$ are stored in the 2D array $scores$, while values of $J$ are stored in the 2D array $mem$.

        \begin{algorithm*}
\caption{DP Algorithm}\label{alg:dp}
\begin{algorithmic}[1]
\State $mem[L][L] \text{ initialized with } \mathtt{INF}$
\State $indices[L][L] \text{ initialized with }-1$
\For{$i = L-1 \textbf{ down to } 0$}
    \For{$j = i+1 : L-1$}
        \State $mem[i, j] \gets scores[i, j]$
        \State $indices[i][j] \gets j$
        \For{k = i+1 : j-1}
            \If{$scores[i][k] + mem[k][j] < mem[i, j]$}
                \State $mem[i, j] \gets scores[i][k] + mem[k][j]$
                \State $indices[i][j] \gets k$
            \EndIf
        \EndFor
    \EndFor
\EndFor

\State $t \gets 0$
\State $Thresholds \gets []$
\While {$t \neq -1$}
    \State $t \gets indices[t][L-1]$
    \State $Thresholds.push(t)$
\EndWhile

\end{algorithmic}
\end{algorithm*}

\begin{algorithm*}
\caption{DP Algorithm with a memory optimization}\label{alg:dp_optmem}
\begin{algorithmic}[1]
\State $mem[L][L] \text{ initialized with } \mathtt{INF}$
\State $next[L] \text{ initialized with }-1$
\For{$i = L-1 \textbf{ down to } 0$}
    \For{$j = i+1 : L-1$}
        \State $mem[i, j] \gets scores[i, j]$
        \State $next[i] \gets j$
        \For{k = i+1 : j-1}
            \If{$scores[i][k] + mem[k][j] < mem[i, j]$}
                \State $mem[i, j] \gets scores[i][k] + mem[k][j]$
                \State $next[i] \gets k$
            \EndIf
        \EndFor
    \EndFor
\EndFor

\State $t \gets 0$
\State $Thresholds \gets []$
\While {$t \neq -1$}
    \State $t \gets next[t]$
    \State $Thresholds.push(t)$
\EndWhile

\end{algorithmic}
\end{algorithm*}

\begin{align}
    \label{eqn:dp-recurrence}
    & J(i, j) = \notag \\
    & \best \left ( Q(i, j), \best_{k=i+1}^{j-1}{\{Q(i,k)+J(k,j)\} } \right )
\end{align}

\subsubsection{Time Complexity Analysis}
\label{subsubsec:time-complexity}

Lines 1-2 in \Cref{alg:dp} are just initialization steps. We have then a loop over $L$ intensity levels. Inside it, there is a nested loop for $(L-1)-(i+1)+1=L-1-i$ iterations. For simplicity, the former can be relaxed as $L$ iterations. Then in lines 7-12, we have a for loop for $(j-1)-(i+1)+1 = j-i-1$ iterations.
Since the lowest value for $i$ is $0$ and the highest value for $j$ is $L-1$, then the highest value for $j-i$ is $(L-1)-(0)=L-1$. Hence, we can relax the $j-i-1$ iterations to be $L$ iterations. Therefore, the worst case complexity for lines 3-14 are $O(L \cdot L \cdot(2+ L \cdot (3))) = O(2L^2+3L^3)=O(L^3)$.
Lines 15 and 16 just set values. The while loop in lines 17-20 traces the best found set of thresholds where they are chained together by the array $next$. At worst case, all $L$ intensity levels will be selected as thresholds. In that case, worst case complexity of lines 17-20 are $O(L \cdot 2) = O(L)$.
Now to sum up, the overall complexity of \Cref{alg:dp} is $O(2+L^3+2+L)=O(L^3)$.

\subsubsection{Space Complexity Analysis}
The original \Cref{alg:dp} has two 2D arrays $mem$ and $indices$ where each one of them contains $L \times L$ floats and integers, respectively. It also assumes the 2D array $scores$ contains precomputed \ac{met} class cost values. $scores$ is a 2D array of $L \times L$ floats. The algorithm utilizes three counter integer variables $i,j,k$ at maximum in Lines 3-14. In the remaining part (Lines 15-17), one integer variable $t$ is utilized in addition to the list $Thresholds$. The list $Thresholds$ starts empty and start to grow with each found threshold. At worst case, all intensity levels of the histogram may be found as proper thresholds. In such case, the array $Thresholds$ will have, at maximum, $L$ elements. The overall space complexity of the \cref{alg:dp} is $O(L^2+L^2+L^2+3 + 1 + L) = O(3L^2 + L + 4) = O(L^2)$.

For the improved variant, \cref{alg:dp_optmem}, it differs from \cref{alg:dp} in the array $indices$. Array $indices$ in \cref{alg:dp_optmem} is a 1D array of $L$ elements instead of a 2D array of $L^2$ elements in \cref{alg:dp}. Hence, space complexity of \cref{alg:dp_optmem} is $O(L^2 +L + L^2 + 3 + 1 + L) = O(2L^2 + L + 4) = O(L^2)$.

\subsubsection{Modified \acf{met} criterion}
    \label{subsubsec:mod-met-criterion}

Traditionally, a threshold $T_i$ separates all pixels having intensity levels below $T_i$ from all of these having intensity levels above $T_i$. For those pixels having the exact intensity level $T_i$, the choice of whether they belong to the lower or higher class is arbitrary.
In other words, a thresholding solution produced by a thresholding scheme that includes the threshold intensity level with the lower class has an equivalent thresholding solution produced by a thresholding scheme including the threshold intensity level with the higher class. This equivalence is trivially constructable my mapping $\{T_{a_1}, ...,T_{a_m}\}$ to $\{T_{a_1}+1, ..., T_{a_m}+1 \}$.
The equivalence between these two solutions in these two schemes is in the sense that thresholding a specific input image by the first solution with its thresholding scheme produces the same exact output that is produced by applying the second thresholding solution with its thresholding scheme.

As in \cref{eqn:generic-eval-fun}, the threshold value $T_{i}$ is considered a member of only the {(i+1)}\textsuperscript{th} class only. In some works, $T_{i}$ is considered a member of only the i\textsuperscript{th} class. This happens by having $J(T_1, ..., T_n) = \sum_{i=1}^{n+1}{Q(T_{i-1}+1, T_{i})}$ while $T_0$ and $T_{n+1}$ are defined to be $-1$ and $L-1$ instead of $0$ and $L$. Both of these are equivalent in the sense that they use different sets of thresholds to refer to the same split of histogram.

This modification might seem to be equivalent to both of the aforementioned approaches; however, it is not. Equivalence here is defined in the sense that minimizing the modified criterion function $J'$ does not necessarily yield the same set of thresholds as minimizing the original criterion $J$.
When applying the resulting thresholding solution to the input image, we keep the same conventional way of including $T_i$ with the (i+1)\textsuperscript{th} class as we cannot include a pixel into two classes simultaneously. Thus, this modification affects the values of the criterion function, and thereby the chosen set of thresholds, but it applies a set of thresholds in the same conventional way of including $T_i$ with only the (i+1)\textsuperscript{th} class.

The function $J'$, representing the value of the modified \ac{met} criterion is calculated according to the following equation.

\begin{equation}f
    J'(\{T_1,..,T_i,..,T_n\})=\sum_{i=1}^{n+1}{w_i\cdot (\ln{\sigma_i}-\ln{w_i})}
    \label{met_mod_eq_original}
\end{equation}\\

where $w_i$ and $\sigma_i$ are calculated according to \cref{eq:weight-interval,eq:std-interval} with $[a,b]=[T_{i-1},T_i]$. Note that the distinction between (\cref{met_mod_eq_original}) and (\cref{eqn:generic-eval-fun,eqn:kittler-q}) is that the terms $w_i$ and $\sigma_i$ are calculated based on the intervals $[T_{i-1},T_i]$ and $[T_{i-1},T_{i}-1]$, respectively.

\begin{figure*}[]
    \centering
    \begin{tikzpicture}
        \node[inner sep=0pt] (img1) at (0,0) {\includegraphics[width=0.48\textwidth]{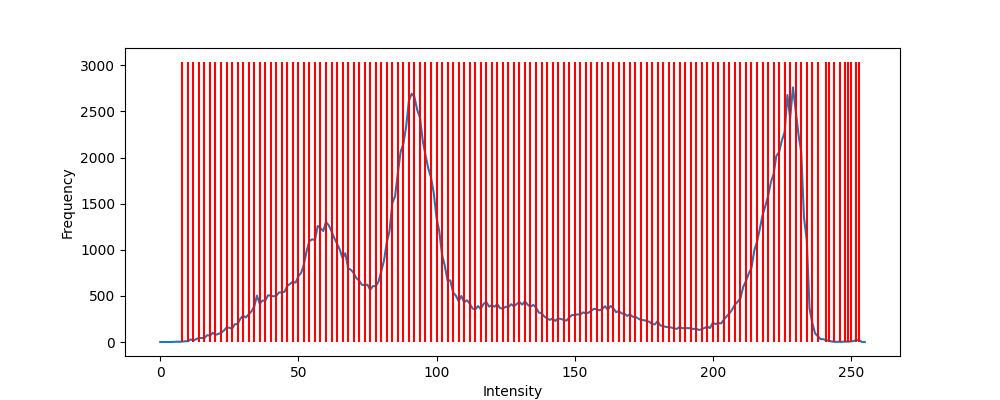}};
        \node[inner sep=0pt] (img2) at (8,0) {\includegraphics[width=0.48\textwidth]{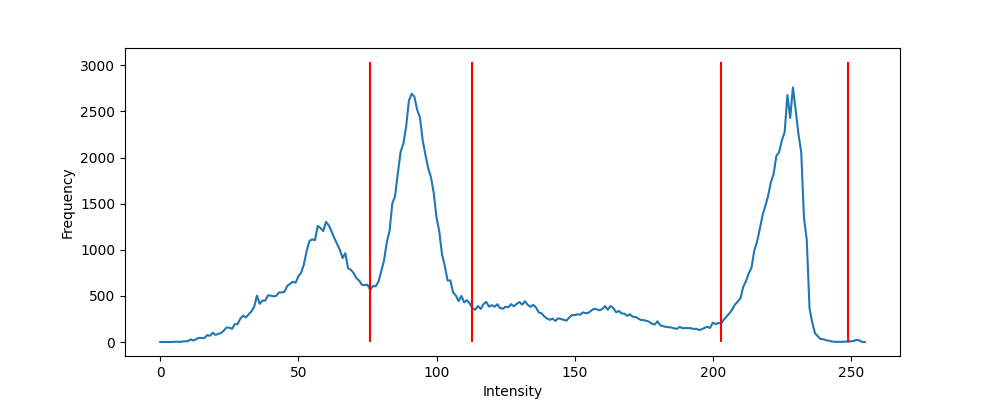}};
        \draw[->, ultra thick, >=stealth] (img1.east) -- (img2.west);
    \end{tikzpicture}

    \caption{Illustration showing effects of modifying \ac{met} criterion used by the proposed dynamic programming variant. Shown on the left is the histogram with thresholds outputted by MET-DP when using the original \ac{met} criterion (disjoint). Shown on the right is the histogram with thresholds outputted by MET-DP after modifying \ac{met} criterion (overlapping). The histogram is for image 46076 from train set of \ac{bsds500}.}
    \label{fig:image_arrow}
\end{figure*}

\subsubsection{Illustrative Example}

\begin{figure}
    \centering
    \includegraphics[width=\linewidth]{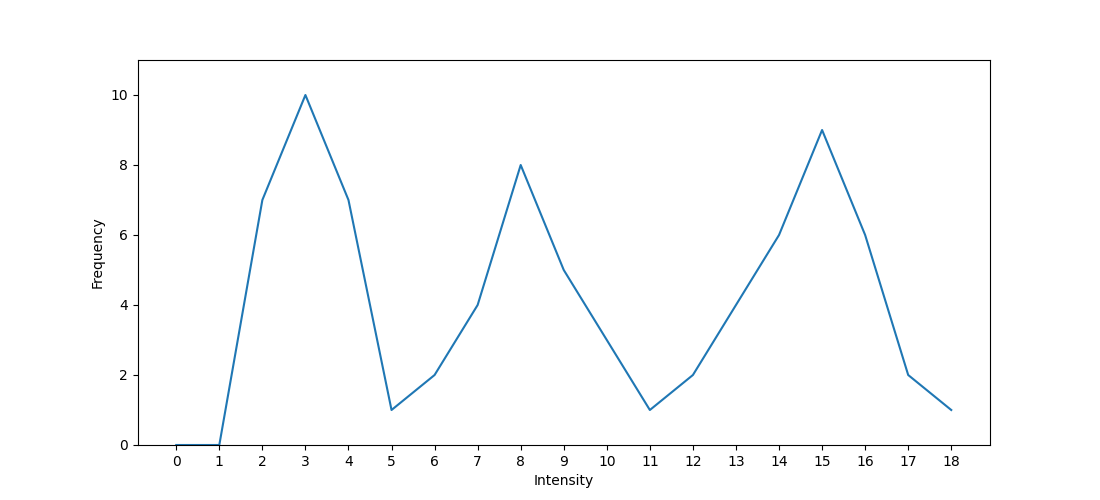}
    \caption{Plot of the histogram used for illustrating the improvement caused by \cref{alg:dp_optmem}. Histogram numerical data is shown in \cref{tab:met-mod-example-plot}. Intensity Levels in this example are limited to $\{0,1,...,18\}$}.
    \label{fig:met-mod-example-plot}
\end{figure}

To illustrate how the modification of \ac{met} criterion improves the results of \cref{alg:dp_optmem}, we show a simplified synthesized example. We assume a histogram of the 19 unique values of $0, 1, ..., 17, 18$. The histogram data is listed in \cref{tab:met-mod-example-plot}. The header row shows indices or unique values. The row with header $h(i)$ shows the frequency of the value $i$. The row with header $i \cdot h(i)$ lists the corresponding value of multiplying the value $i$ by the count of its occurrences (its frequency). This value $i \cdot h(i)$ is used in computation of $\mu$ of a class, as in \cref{eq:mean-interval}. The row with header $i \cdot i \cdot h(i)$ shows the values of multiplying $i^2$ with the count of occurrences of $i$. This value is used in computation of $\sigma^2$ of a class, as in \cref{eq:variance-interval}.
We also show in this table the values of cumulative summation of each of these three expressions. Cumulative summation is used to optimize the calculation of summation of any of these three expressions ($h(i)$, $i\cdot h(i)$ and $i \cdot i \cdot h(i)$) over any subinterval of the full interval $[0,18]$. 
Looking at the histogram plot at \cref{fig:met-mod-example-plot}, we can see this simple histogram is tri-modal. There is one distribution in $[1,5]$, one in $[6,10]$ and one at $[11,16]$. Hence, it is apparent that for such a simple histogram, the best number of thresholds is 2 thresholds with each of them lying somewhere in $[5,6]$ and $[10,11]$, respectively. 
Based on the values reported in \cref{tab:met-mod-example-plot}, the class score value is calculated for each possible class with bounds $[a,b]$ where $0 \leq a<b\leq 19-1$. The class score function is computed according to \cref{eq:class-cost}, taken from \cref{met_mod_eq_original}. The computed values for every possible class are reported in \cref{tab:met_dp_example_scores}. The cells on and below the main diagonal are empty as they represent empty intervals $[a,b]$ with $a \geq b$. The singleton classes $[a,a]$ are ignored due to their insignificance. 

\begin{equation}
    Q_i=w_i \cdot \left ( \ln{\sigma_i} - \ln{w_i} \right )
    \label{eq:class-cost}
\end{equation}
where $w_i$ and $\sigma_i$ are calculated based on \cref{eq:weight-interval,eq:std-interval,eq:variance-interval} with intervals $[T_{i-1},T_{i}]$

\renewcommand{\arraystretch}{1.5}
\begin{table*}[]
    \centering
    \begin{adjustbox}{max width=\textwidth}
    \begin{tabular}{c|lllllllllllllllllll}
         i &  0 &   1 &   2 &   3 &   4 &   5 &   6 &   7 &   8 &   9 &  10 &  11 &  12 &  13 &  14 &  15 &  16 &  17 &  18 \\ \midrule
         $h(i)$ & 0 &   0 &   7 &  10 &   7 &   1 &   2 &   4 &   8 &   5 &   3 &   1 &   2 &   4 &   6 &   9 &   6 &   2 &   1 \\
         $i \cdot h(i)$ &  0 &   0 &  14 &  30 &  28 &   5 &  12 &  28 &  64 &  45 &  30 &  11 &  24 &  52 &  84 &  135 &  96 &  34 &  18 \\
         $i \cdot i \cdot h(i)$ &   0 &   0 &  28 &  90 &  112 &  25 &  72 &  196 &  512 &  405 &  300 &  121 &  288 &  676 &  1176 &  2025 &  1536 &  578 &  324 \\
         $\sum_{j=0}^{i}{(h(j))}$ & 0 &   0 &   7 &  17 &  24 &  25 &  27 &  31 &  39 &  44 &  47 &  48 &  50 &  54 &  60 &  69 &  75 &  77 &  78 \\
        $\sum_{j=0}^{i}{(j \cdot h(j))}$ & 0 &   0 &  14 &  44 &  72 &  77 &  89 &  117 &  181 &  226 &  256 &  267 &  291 &  343 &  427 &  562 &  658 &  692 &  710 \\
        $\sum_{j=0}^{i}{(j \cdot j \cdot h(j))}$ &0 &   0 &  28 &  118 &  230 &  255 &  327 &  523 &  1035 &  1440 &  1740 &  1861 &  2149 &  2825 &  4001 &  6026 &  7562 &  8140 &  8464 \\
    \end{tabular}
    \end{adjustbox}
    \caption{Synthesized histogram for illustrating the improvement caused by \cref{alg:dp_optmem}. Histogram is shown in \cref{fig:met-mod-example-plot}.  Intensity Levels in this example are limited to $\{0,1,...,18\}$. This table shows values of the histogram $h(i)$, first moment of the histogram $i \cdot h(i)$, the second moment of the histogram $i \cdot i \cdot h(i)$ and the cumulative summations starting from $0$ up to $i$ inclusively for each of these three quantities. These values are used in computation of $Q(i,j)$ for different thresholding objective functions as explained in \cref{subsec:thr-prob-form}.}
    \label{tab:met-mod-example-plot}
\end{table*}
\renewcommand{\arraystretch}{1}

\begin{figure}[]
    \centering
    \begin{tikzpicture}
        \node[inner sep=0pt] (img1) at (0,5) {\includegraphics[width=0.48\textwidth]{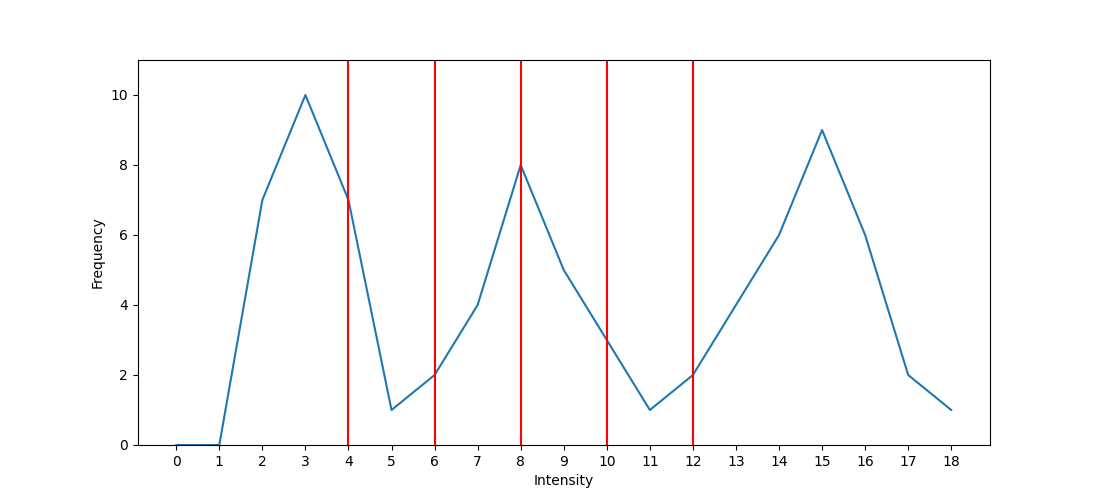}};
        \node[inner sep=0pt] (img2) at (0,0) {\includegraphics[width=0.48\textwidth]{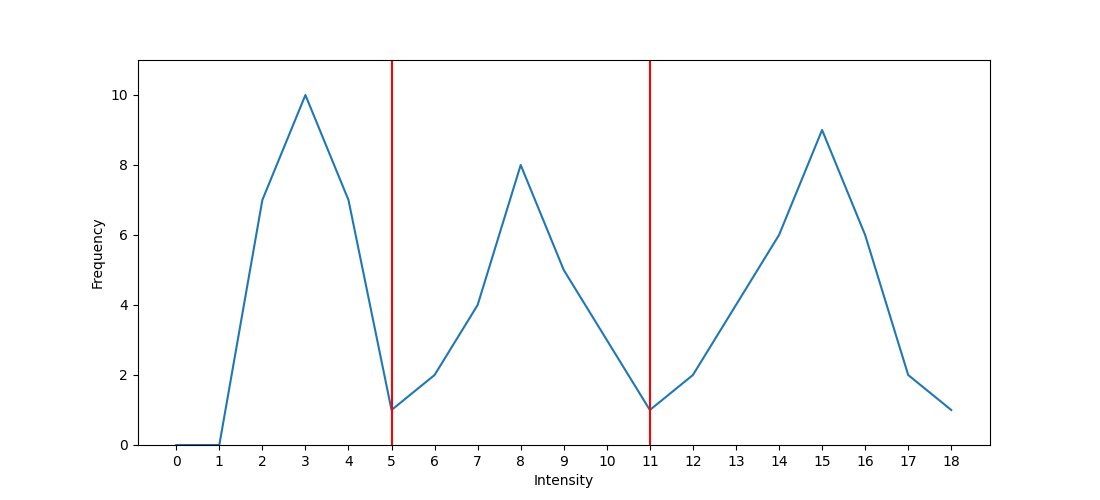}};
        \draw[->, ultra thick, >=stealth] (img1.south) -- (img2.north);
    \end{tikzpicture}

    \caption{
        Illustration showing effects of modifying \ac{met} criterion used by the proposed dynamic programming variant. Shown on top is the histogram with thresholds outputted when using the original criterion while on bottom is the histogram with thresholds outputted after modifying the \ac{met} criterion. The histogram is for the artificial example detailed in \cref{fig:met-mod-example-plot,tab:met-mod-example-plot}.
    }
    \label{fig:met-mod-example-disj-arrow-over}
\end{figure}

\renewcommand{\arraystretch}{1.7}
\begin{sidewaystable*}[]
\centering
\begin{adjustbox}{max width=\textwidth}
    \begin{tabular}{||r||R|R|R|R|R|R|R|R|R|R|R|R|R|R|R|R|R|R|R||}
    \toprule
    \multicolumn{1}{||r||}{$a/b$} & \multicolumn{1}{|r|}{0} & \multicolumn{1}{|r|}{1} & \multicolumn{1}{|r|}{2} & \multicolumn{1}{|r|}{3} & \multicolumn{1}{|r|}{4} & \multicolumn{1}{|r|}{5} & \multicolumn{1}{|r|}{6} & \multicolumn{1}{|r|}{7} & \multicolumn{1}{|r|}{8} & \multicolumn{1}{|r|}{9} & \multicolumn{1}{|r|}{10} & \multicolumn{1}{|r|}{11} & \multicolumn{1}{|r|}{12} & \multicolumn{1}{|r|}{13} & \multicolumn{1}{|r|}{14} & \multicolumn{1}{|r|}{15} & \multicolumn{1}{|r|}{16} & \multicolumn{1}{|r|}{17} & \multicolumn{1}{|r||}{18} \\
    \midrule
    0  & & 0.000 & 0.000 & 0.256 & 0.404 & 0.448 & 0.585 & 0.806 & 1.080 & 1.218 & 1.308 & 1.345 & 1.433 & 1.607 & 1.831 & 2.099 & 2.255 & 2.309 & 2.341 \\ 
    \hline
    1  & &       & 0.000 & 0.256 & 0.404 & 0.448 & 0.585 & 0.806 & 1.080 & 1.218 & 1.308 & 1.345 & 1.433 & 1.607 & 1.831 & 2.099 & 2.255 & 2.309 & 2.341 \\
    \hline
    2  & &       &        & 0.256 & 0.404 & 0.448 & 0.585 & 0.806 & 1.080 & 1.218 & 1.308 & 1.345 & 1.433 & 1.607 & 1.831 & 2.099 & 2.255 & 2.309 & 2.341 \\
    \hline
    3  & &       &        &        & 0.256 & 0.319 & 0.481 & 0.699 & 0.954 & 1.083 & 1.169 & 1.205 & 1.292 & 1.463 & 1.681 & 1.940 & 2.090 & 2.143 & 2.174 \\
    \hline
    4  & &       &        &        &        & 0.173 & 0.340 & 0.516 & 0.732 & 0.844 & 0.924 & 0.959 & 1.047 & 1.216 & 1.426 & 1.672 & 1.814 & 1.865 & 1.896 \\
    \hline
    5  & &       &        &        &        &        & 0.139 & 0.271 & 0.437 & 0.539 & 0.626 & 0.671 & 0.781 & 0.971 & 1.186 & 1.426 & 1.562 & 1.613 & 1.644 \\
    \hline
    6  & &       &        &        &        &        &        & 0.201 & 0.363 & 0.472 & 0.568 & 0.617 & 0.737 & 0.933 & 1.150 & 1.389 & 1.525 & 1.575 & 1.607 \\
    \hline
    7  & &       &        &        &        &        &        &        & 0.249 & 0.378 & 0.490 & 0.545 & 0.676 & 0.877 & 1.093 & 1.328 & 1.462 & 1.512 & 1.544 \\
    \hline
    8  & &       &        &        &        &        &        &        &        & 0.258 & 0.391 & 0.454 & 0.591 & 0.788 & 0.996 & 1.222 & 1.351 & 1.400 & 1.431 \\
    \hline
    9  & &       &        &        &        &        &        &        &        &        & 0.230 & 0.296 & 0.423 & 0.595 & 0.780 & 0.981 & 1.097 & 1.143 & 1.175 \\
    \hline
    10 & &       &        &        &        &        &        &        &        &        &        & 0.158 & 0.273 & 0.424 & 0.589 & 0.765 & 0.870 & 0.918 & 0.954 \\ \hline
    11 & &       &        &        &        &        &        &        &        &        &        &        & 0.139 & 0.271 & 0.418 & 0.577 & 0.683 & 0.738 & 0.783 \\ \hline
    12 & &       &        &        &        &        &        &        &        &        &        &        &        & 0.201 & 0.350 & 0.509 & 0.620 & 0.680 & 0.729 \\ \hline
    13 & &       &        &        &        &        &        &        &        &        &        &        &        &        & 0.248 & 0.411 & 0.530 & 0.597 & 0.653 \\ \hline
    14 & &       &        &        &        &        &        &        &        &        &        &        &        &        &        & 0.259 & 0.401 & 0.482 & 0.549 \\ \hline
    15 & &       &        &        &        &        &        &        &        &        &        &        &        &        &        &        & 0.259 & 0.363 & 0.442 \\ \hline
    16 & &       &        &        &        &        &        &        &        &        &        &        &        &        &        &        &        & 0.213 & 0.296 \\ \hline
    17 & &       &        &        &        &        &        &        &        &        &        &        &        &        &        &        &        &        & 0.139 \\ \hline
    18 & &       &        &        &        &        &        &        &        &        &        &        &        &        &        &        &        &        &       \\
    \bottomrule
    \end{tabular}

\end{adjustbox}
\caption{Values of $Q(a,b)$ for the illustrative example detailed in \cref{fig:met-mod-example-plot,tab:met-mod-example-plot}. $Q(a,b)$ is the total error value for the histogram region $[a, b]$ and was calculated according to \cref{eqn:kittler-q}. }
\label{tab:met_dp_example_scores}
\end{sidewaystable*}
\renewcommand{\arraystretch}{1}

\renewcommand{\arraystretch}{1.5}
\begin{sidewaystable*}[]
\centering
\begin{adjustbox}{max width=\textwidth}
    \begin{tabular}{||r||r|r|r|r|r|r|r|r|r|r|r|r|r|r|r|r|r|r|r||}
    \toprule
    \(i\!/\!j\) & 0 & 1 & 2 & 3 & 4 & 5 & 6 & 7 & 8 & 9 & 10 & 11 & 12 & 13 & 14 & 15 & 16 & 17 & 18 \\
\hline
0 & & \text{INF[1]} & \text{INF[2]} & 0.256[3] & 0.404[4] & 0.448[5] & \cellcolor{cyan} 0.585[6] & \cellcolor{lightgray} 0.719[5] & \cellcolor{lightgray} 0.885[5] & \cellcolor{lightgray} 0.987[5] & \cellcolor{lightgray} 1.074[5] & \cellcolor{lightgray} 1.119[5] & \cellcolor{lightgray} 1.229[5] & \cellcolor{lightgray} 1.390[5] & \cellcolor{lightgray} 1.537[5] & \cellcolor{lightgray} 1.696[5] & \cellcolor{lightgray} 1.802[5] & \cellcolor{lightgray} 1.857[5] & \cellcolor{lightgray} 1.901[5] \\
1 & & & \text{INF[2]} & 0.256[3] & 0.404[4] & 0.448[5] & \cellcolor{cyan}  0.585[6] & \cellcolor{lightgray} 0.719[5] & \cellcolor{lightgray} 0.885[5] & \cellcolor{lightgray} 0.987[5] & \cellcolor{lightgray} 1.074[5] & \cellcolor{lightgray} 1.119[5] & \cellcolor{lightgray} 1.229[5] & \cellcolor{lightgray} 1.390[5] & \cellcolor{lightgray} 1.537[5] & \cellcolor{lightgray} 1.696[5] & \cellcolor{lightgray} 1.802[5] & \cellcolor{lightgray} 1.857[5] & \cellcolor{lightgray} 1.901[5] \\
2 & & & & 0.256[3] & 0.404[4] & 0.448[5] & \cellcolor{cyan} 0.585[6] & \cellcolor{lightgray} 0.719[5] & \cellcolor{lightgray} 0.885[5] & \cellcolor{lightgray} 0.987[5] & \cellcolor{lightgray} 1.074[5] & \cellcolor{lightgray} 1.119[5] & \cellcolor{lightgray} 1.229[5] & \cellcolor{lightgray} 1.390[5] & \cellcolor{lightgray} 1.537[5] & \cellcolor{lightgray} 1.696[5] & \cellcolor{lightgray} 1.802[5] & \cellcolor{lightgray} 1.857[5] & \cellcolor{lightgray} 1.901[5] \\
3 & & & & & 0.256[4] & 0.319[5] & \cellcolor{lightgray} 0.458[5] & \cellcolor{lightgray} 0.590[5] & \cellcolor{lightgray} 0.755[5] & \cellcolor{lightgray} 0.857[5] & \cellcolor{lightgray} 0.945[5] & \cellcolor{lightgray} 0.989[5] & \cellcolor{lightgray} 1.100[5] & \cellcolor{lightgray} 1.260[5] & \cellcolor{lightgray} 1.407[5] & \cellcolor{lightgray} 1.566[5] & \cellcolor{lightgray} 1.672[5] & \cellcolor{lightgray} 1.727[5] & \cellcolor{lightgray} 1.772[5] \\
4 & & & & & & 0.173[5] & \cellcolor{lightgray} 0.312[5] & \cellcolor{lightgray} 0.444[5] & \cellcolor{lightgray} 0.610[5] & \cellcolor{lightgray} 0.712[5] & \cellcolor{lightgray} 0.800[5] & \cellcolor{lightgray} 0.844[5] & \cellcolor{lightgray} 0.954[5] & \cellcolor{lightgray} 1.115[5] & \cellcolor{lightgray} 1.262[5] & \cellcolor{lightgray} 1.421[5] & \cellcolor{lightgray} 1.527[5] & \cellcolor{lightgray} 1.582[5] & \cellcolor{lightgray} 1.627[5] \\
5 & & & & & & & 0.139[6] & 0.271[7] & 0.437[8] & 0.539[9] & 0.626[10] & 0.671[11] & 0.781[12] & \cellcolor{gray} 0.942[11] &\cellcolor{gray}  1.089[11] & \cellcolor{gray} 1.247[11] & \cellcolor{gray} 1.354[11] & \cellcolor{gray} 1.409[11] & \cellcolor{gray} 1.453[11] \\
6 & & & & & & & & 0.201[7] & 0.363[8] & 0.472[9] & 0.568[10] & 0.617[11] & 0.737[12] & \cellcolor{gray} 0.888[11] & \cellcolor{gray} 1.035[11] & \cellcolor{gray} 1.194[11] & \cellcolor{gray} 1.300[11] & \cellcolor{gray} 1.355[11] & \cellcolor{gray} 1.400[11] \\
7 & & & & & & & & & 0.249[8] & 0.378[9] & 0.490[10] & 0.545[11] & 0.676[12] & \cellcolor{gray} 0.816[11] & \cellcolor{gray} 0.963[11] & \cellcolor{gray} 1.122[11] & \cellcolor{gray} 1.228[11] & \cellcolor{gray} 1.283[11] & \cellcolor{gray} 1.328[11] \\
8 & & & & & & & & & & 0.258[9] & 0.391[10] & 0.454[11] & 0.591[12] & \cellcolor{gray} 0.725[11] & \cellcolor{gray} 0.872[11] & \cellcolor{gray} 1.031[11] & \cellcolor{gray} 1.137[11] & \cellcolor{gray} 1.192[11] & \cellcolor{gray} 1.236[11] \\
9 & & & & & & & & & & & 0.230[10] & 0.296[11] & 0.423[12] & \cellcolor{gray} 0.568[11] & \cellcolor{gray} 0.715[11] & \cellcolor{gray} 0.873[11] & \cellcolor{gray} 0.979[11] & \cellcolor{gray} 1.035[11] & \cellcolor{gray} 1.079[11] \\
10 & & & & & & & & & & & & 0.158[11] & 0.273[12] & 0.424[13] & \cellcolor{gray}0.576[11] &\cellcolor{gray} 0.735[11] & \cellcolor{gray}0.841[11] &\cellcolor{gray} 0.896[11] & \cellcolor{gray} 0.941[11] \\
11 & & & & & & & & & & & & & 0.139[12] & 0.271[13] & 0.418[14] & 0.577[15] & 0.683[16] & 0.738[17] & 0.783[18] \\
12 & & & & & & & & & & & & & & 0.201[13] & 0.350[14] & 0.509[15] & 0.620[16] & 0.680[17] & 0.729[18] \\
13 & & & & & & & & & & & & & & & 0.248[14] & 0.411[15] & 0.530[16] & 0.597[17] & 0.653[18] \\
14 & & & & & & & & & & & & & & & & 0.259[15] & 0.401[16] & 0.482[17] & 0.549[18] \\
15 & & & & & & & & & & & & & & & & & 0.259[16] & 0.363[17] & 0.442[18] \\
16 & & & & & & & & & & & & & & & & & & 0.213[17] & 0.296[18] \\
17 & & & & & & & & & & & & & & & & & & & 0.139[18] \\
18 & & & & & & & & & & & & & & & & & & & \\
    \bottomrule
    \end{tabular}

\end{adjustbox}
\caption{Values of $mem(i,j)$ and $indices[i,j]$ for an illustrative example ; using MET-DP Overlapping}
\label{tab:met_dp_example_mem_ind}
\end{sidewaystable*}
\renewcommand{\arraystretch}{1}

\renewcommand{\arraystretch}{1.5}
\begin{sidewaystable*}[]
\centering
\begin{adjustbox}{max width=\textwidth}
    \begin{tabular}{||r||r|r|r|r|r|r|r|r|r|r|r|r|r|r|r|r|r|r|r||}
    \toprule
    \(i\!/\!j\) & 0 & 1 & 2 & 3 & 4 & 5 & 6 & 7 & 8 & 9 & 10 & 11 & 12 & 13 & 14 & 15 & 16 & 17 & 18 \\
\hline
0  &        & INF[1] & INF[2] & 0.256[3] & 0.404[4] & 0.429[4] & 0.543[5] & 0.631[4] & 0.791[5] & 0.888[4] & 0.998[4] & 1.046[4] & 1.137[4] & 1.247[4] & 1.385[4] & 1.507[4] & 1.644[4] & 1.720[4] & 1.775[4] \\
1  &        &         & INF[2] & 0.256[3] & 0.404[4] & 0.429[4] & 0.543[5] & 0.631[4] & 0.791[5] & 0.888[4] & 0.998[4] & 1.046[4] & 1.137[4] & 1.247[4] & 1.385[4] & 1.507[4] & 1.644[4] & 1.720[4] & 1.775[4] \\
2  &        &         &         & 0.256[3] & 0.404[4] & 0.429[4] & 0.543[5] & 0.631[4] & 0.791[5] & 0.888[4] & 0.998[4] & 1.046[4] & 1.137[4] & 1.247[4] & 1.385[4] & 1.507[4] & 1.644[4] & 1.720[4] & 1.775[4] \\
3  &        &         &         &         & 0.256[4] & 0.319[5] & 0.395[5] & 0.520[6] & 0.644[5] & 0.773[5] & 0.873[5] & 0.927[5] & 1.012[5] & 1.128[5] & 1.260[5] & 1.387[5] & 1.520[5] & 1.600[5] & 1.655[5] \\
4  &        &         &         &         &         & 0.173[5] & 0.340[6] & 0.374[6] & 0.536[6] & 0.632[6] & 0.742[6] & 0.790[6] & 0.881[6] & 0.991[6] & 1.129[6] & 1.250[6] & 1.388[6] & 1.464[6] & 1.519[6] \\
5  &        &         &         &         &         &         & 0.139[6] & 0.271[7] & 0.388[7] & 0.517[7] & 0.617[7] & 0.671[11] & 0.756[7] & 0.872[12] & 1.004[7] & 1.131[12] & 1.264[7] & 1.344[12] & 1.399[12] \\
6  &        &         &         &         &         &         &         & 0.201[7] & 0.363[8] & 0.459[8] & 0.568[10] & 0.617[8] & 0.707[11] & 0.818[8] & 0.955[11] & 1.077[8] & 1.215[11] & 1.290[8] & 1.345[8] \\
7  &        &         &         &         &         &         &         &         & 0.249[8] & 0.378[9] & 0.478[9] & 0.536[10] & 0.617[9] & 0.737[10] & 0.865[9] & 0.996[10] & 1.125[9] & 1.210[10] & 1.261[9] \\
8  &        &         &         &         &         &         &         &         &         & 0.258[9] & 0.391[10] & 0.415[10] & 0.530[11] & 0.617[10] & 0.766[10] & 0.876[10] & 1.018[10] & 1.089[10] & 1.144[10] \\
9  &        &         &         &         &         &         &         &         &         &         & 0.230[10] & 0.296[11] & 0.369[11] & 0.498[12] & 0.617[11] & 0.757[12] & 0.876[11] & 0.966[11] & 1.012[11] \\
10 &        &         &         &         &         &         &         &         &         &         &         & 0.158[11] & 0.273[12] & 0.359[12] & 0.508[12] & 0.618[12] & 0.760[12] & 0.832[12] & 0.887[12] \\
11 &        &         &         &         &         &         &         &         &         &         &         &         & 0.139[12] & 0.271[13] & 0.387[13] & 0.531[14] & 0.646[13] & 0.737[13] & 0.783[18] \\
12 &        &         &         &         &         &         &         &         &         &         &         &         &         & 0.201[13] & 0.350[14] & 0.461[14] & 0.602[14] & 0.674[14] & 0.729[18] \\
13 &        &         &         &         &         &         &         &         &         &         &         &         &         &         & 0.248[14] & 0.411[15] & 0.507[15] & 0.597[17] & 0.646[15] \\
14 &        &         &         &         &         &         &         &         &         &         &         &         &         &         &         & 0.259[15] & 0.401[16] & 0.473[16] & 0.540[17] \\
15 &        &         &         &         &         &         &         &         &         &         &         &         &         &         &         &         & 0.259[16] & 0.363[17] & 0.398[17] \\
16 &        &         &         &         &         &         &         &         &         &         &         &         &         &         &         &         &         & 0.213[17] & 0.296[18] \\
17 &        &         &         &         &         &         &         &         &         &         &         &         &         &         &         &         &         &         & 0.139[18] \\
18 &        &         &         &         &         &         &         &         &         &         &         &         &         &         &         &         &         &         &         \\

    \bottomrule
    \end{tabular}

\end{adjustbox}
\caption{Values of $mem(i,j)$ and $indices[i,j]$ for an illustrative example ; using MET-DP Disjoint}
\label{tab:met_dp_example_mem_ind_disj}
\end{sidewaystable*}
\renewcommand{\arraystretch}{1}

For instance, take cell $[i=8,j=17]$ in \cref{tab:met_dp_example_mem_ind}. $mem[8,17]$ is to be calculated as given by \cref{eqn:dp-recurrence}. In particular, a comparison between the following values are done and the minimum out of them is taken.
\begin{enumerate}
    \item $scores[8,17] = 1.400$
    \item $scores[8,9] + mem[9,17] = 0.258 + 1.035 = 1.292$
    \item $scores[8,10] + mem[10,17] = 0.391 + 0.896 = 1.287$
    \item \label{mink} $scores[8,11] + mem[11,17] = 0.454 + 0.738 = 1.192$
    \item $scores[8,12] + mem[12,17] = 0.591 + 0.680 = 1.271$
    \item $scores[8,13] + mem[13,17] = 0.788 + 0.597 = 1.386$
    \item $scores[8,14] + mem[14,17] = 0.996 + 0.482 = 1.478$
    \item $scores[8,15] + mem[15,17] = 1.222 + 0.363 = 1.585$
    \item $scores[8,16] + mem[16,17] = 1.351 + 0.213 = 1.564$
\end{enumerate}

Note that in the previous calculations, the values are shown after approximation to 3 decimal digits while the actual calculation was performed on the non-approximated values. Meaning that, the addition $0.258 + 1.035$ normally results in $1.293$. However, it is written as $1.292$ above instead. This is because the actual values are $0.25758113833731044 + 1.0345710388880593 = 1.2921521772253697 \approx 1.292$.

We can now see that the minimum value is \cref{mink} at $k=11$. Hence, $mem[8,17]$ is set to $1.192$ and $indices[8,17]=11$. This means that the minimum \ac{met} criterion value for the region $[8,17]$ is $1.292$ and it is obtained by having a class in the region $[8,11]$ and the region $[11,17]$. The region $[11,17]$ may be formed of one or more classes, depending on $indices[11,17]$  that got computed in a previous iteration.

\section{Results}\label{sec2}

\subsection{Setup}

All experiments reported on this article were implemented in Python 3. The experimental setup was an i7-10750H CPU and 16GB of RAM on a Microsoft Windows 10 operating system.

\subsection{Data}
We test the MET-DP method on different types of images in order to evaluate its robustness in different scenarios and for different types of images. We aim to find scenarios where it works well and pinpoint scenarios where it needs improvement or is not suitable. Hence, we form a set of images compromising: natural images, satellite images and medical images. These are the mostly studied types of images for grayscale images multilevel thresholding \citep{Rai_Das_Dhal_2022}. They form a total of $15$ images. \Cref{tab:test-images} shows the breakdown of the chosen images and their source datasets. The source datasets are \ac{bsds500} \citep{bsds500}, Weizmann \citep{Weizmann}, \ac{ssd-uae} \citep{SSD-UAE}, \ac{isic2016} \citep{isic2016} and \ac{brats2020} \citep{brats1}.

\begin{table}
    \centering
    \begin{tabular}{|c|c|c|}
        \toprule
        Type & Dataset & Selected Amount \\
        \midrule
         Natural & \acs{bsds500} &  3 \\
         Natural &  Weizmann & 2 \\
         Medical & \acs{isic2016} & 1 \\
         Medical & \acs{brats2020} & 4 \\
         Satellite & \acs{ssd-uae} & 5 \\
         \bottomrule
    \end{tabular}
    \caption{Details of the set of images used for testing}
    \label{tab:test-images}
\end{table}

\subsection{Evaluation Criteria}
\label{subsec:eval-crit}



In literature, thresholding methods that take the number of thresholds as input get evaluated based on the segmentation quality, computation time and the balance between them. However, the nature of the MET-DP is different as it does not take the number of thresholds as input. Thus, the found number of thresholds is an extra output that needs to be evaluated. Moreover, evaluating the runtime will be done in a slightly different way. Evaluating the thresholding quality will be done in the standard way in literature. These three evaluation criteria are illustrated by the following three points.
\begin{enumerate}
    \item \begin{description}
        \item[Number of thresholds: ] The determined number of thresholds by the MET-DP method needs to be evaluated. Due to the under representation of this factor of thresholding problem in studies, there is not an standardized numerical metric to evaluate the goodness of a found number of thresholds. Therefore, we evaluate the quality of number of thresholds in a qualitative way similar to previous studies on a case-by-case basis \citep{Olmez_Sengur_Koca_Rao_2023,Liang_Cuevas_2013,Dirami_Hammouche_Diaf_Siarry_2013}.
    \end{description}
    \item
        \label{crit:runtime}
    \begin{description}
        \item[Runtime: ] Since the MET-DP method has one less input than other traditional methods (the count of thresholds), it is not fair to compare its runtime with runtime of other traditional methods. For other traditional methods to be used to determine an appropriate number of thresholds, they will be run multiple times with candidate counts of thresholds, and then the best count out of them is taken (based on an arbitrary criterion). We compare the runtime of the MET-DP method with the total runtime of other traditional methods when run with the count of thresholds ranging from 1 to 15. The upper bound of 10 was chosen as it is the maximum count of thresholds outputted by the MET-DP on the set of test images reported in this article. We do not choose a higher upper bound to avoid bias towards the MET-DP method. Also, the MET-DP method is not restricted to only 10 thresholds, as it theoretically can find more thresholds.

        We are comparing the MET-DP method, which has a complexity of $O(L^3)$, with two instances of the
        conventional dynamic programming approach once using Otsu and once using Kapur. This conventional approach is
        known to have a complexity of $O(n \cdot L^2)$ where $n$ is the number of thresholds \citep{lei_dp_masi,
            Merzban_Elbayoumi_2019}. Since we will be running the conventional dynamic programming approach multiple
        times with different values for $n$ (in order to find an appropriate value for $n$ afterwards), the
        complexity of that will be $O(m \cdot n \cdot L^2)$ where $m$ is the number of times to run the conventional
        method. Hence, we can expect that the runtime of the MET-DP approach will be almost constant, since the only
        source of change of its runtime is computing the histogram (which depends on the number of pixels of the
        input image) and other factors related to the used machine CPU and memory constraints. However, the two other
        methods will have an increasing runtime with the increase of $m$ and $n$. Meaning that, running that approach for $n=1$ will have $L^2$ runtime. Running for $n=1$ and $n=2$ will have $L^2 + 2L^2 = 3L^2$ runtime. In general, assuming the $m$ runs are for values of $n$ ranging from $1$ to $m$ inclusively, its runtime is $\sum_{i=1}^{m}{i \cdot L^2} = L^2 \cdot \sum_{i=1}^{m}{i} = L^2 \cdot \frac{m^2+m}{2} = O(m^2 \cdot L^2)$.

    \end{description}
    \item \begin{description}
        \item[Thresholding Quality: ] We use the same criteria used in literature. We pick \ac{ssim} and \ac{psnr} metrics as they are the most two common metrics used to evaluate segmentation quality of thresholding methods \citep{Amiriebrahimabadi_Rouhi_Mansouri_2024}. However, when we compare the MET-DP method with traditional methods that take the number of thresholds as input, we set the number of thresholds for these methods as the count derived by the MET-DP method.
    \end{description}
\end{enumerate}

\subsection{Number of thresholds}

\begin{figure*}

\plothistograms{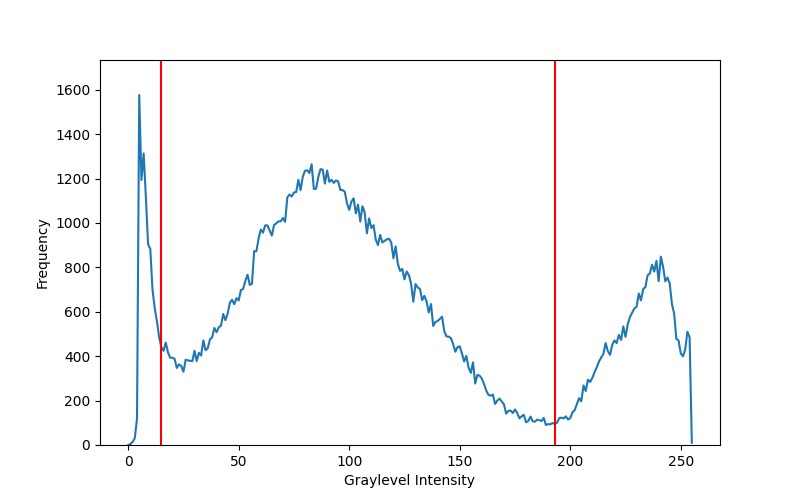}{2}
\plothistograms{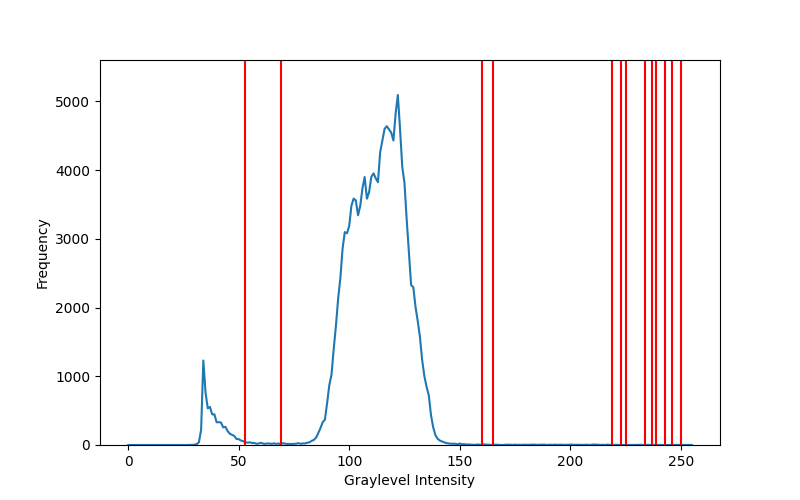}{13}
\plothistograms{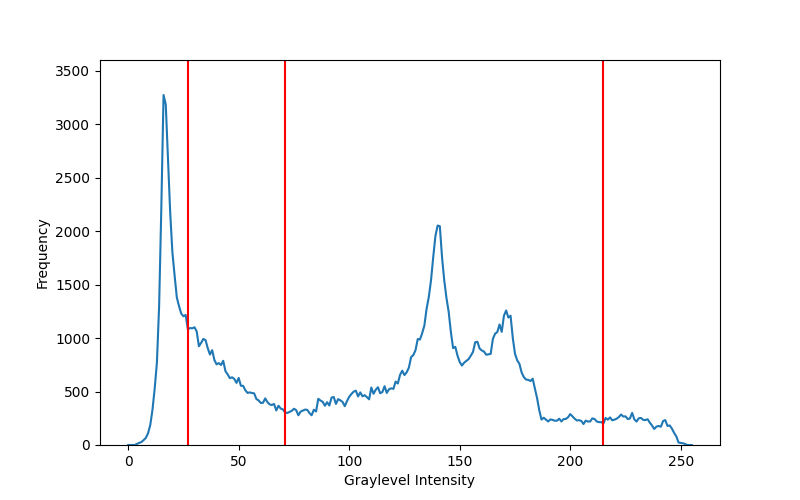}{3}
\plothistograms{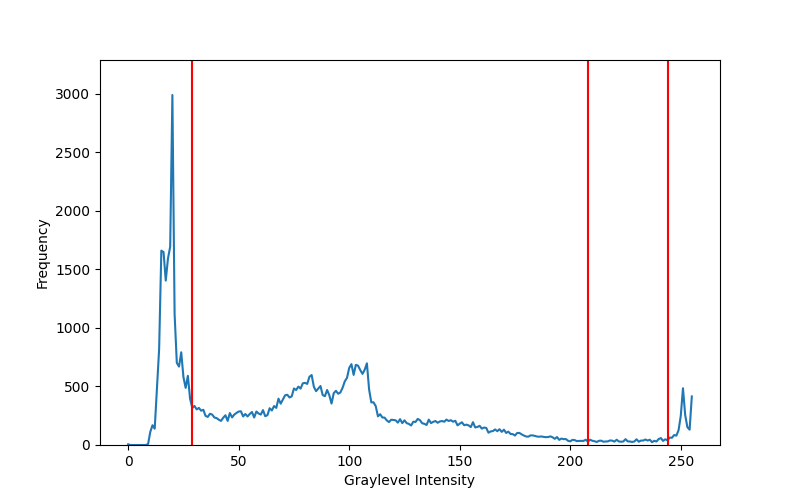}{3}
\plothistograms{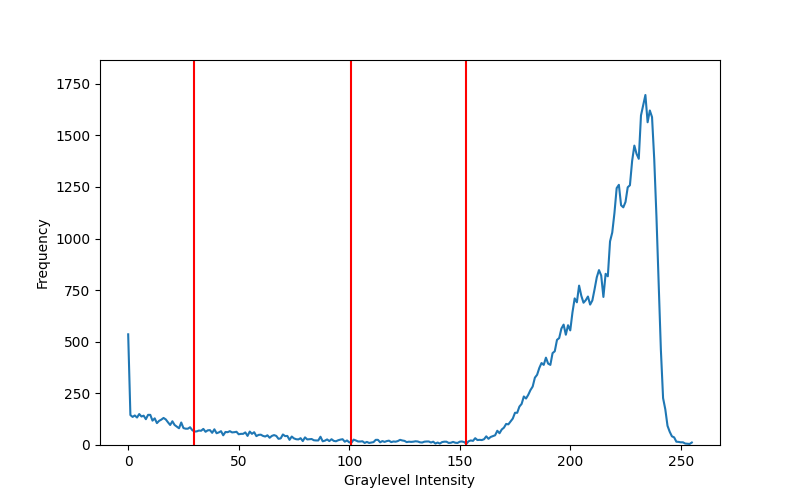}{3}
             \caption{Comparison of DP\_N\_Otsu, DP\_N\_Kapur, DP\_N\_Kittler and MET-DP methods. Each row shows the histogram of a test image with the output thresholds of each methods labeled as vertical red lines. Each column represents the output of one of the four methods. From top to bottom, images \texttt{326085} with $n = 2$, \texttt{135069} with $n = 13$, \texttt{147091} with $n = 3$, \texttt{animal-5-bg-020803} with $n = 3$ and \texttt{77200701} with $n = 3$. From left to right, DP\_N\_Otsu, DP\_N\_Kapur, DP\_N\_Kittler and MET-DP.}
    \label{fig:hs_1}
\end{figure*}

For image $\textsc{326085}$, the histogram in \cref{fig:hs_1} can be clearly seen to be tri-modal. However, Otsu and Kapur fail by using two thresholds to separate these three modes while Kittler and MET-DP succeed. As we can see that  the  part of the second mode is merged with the first mode using Otsu. Similarly, when thresholding with Kapur objective function, the same issue is present. Moreover, part of the second mode is merged with the third mode.

\begin{figure}
    \centering
    \includegraphics[width=0.9\linewidth]{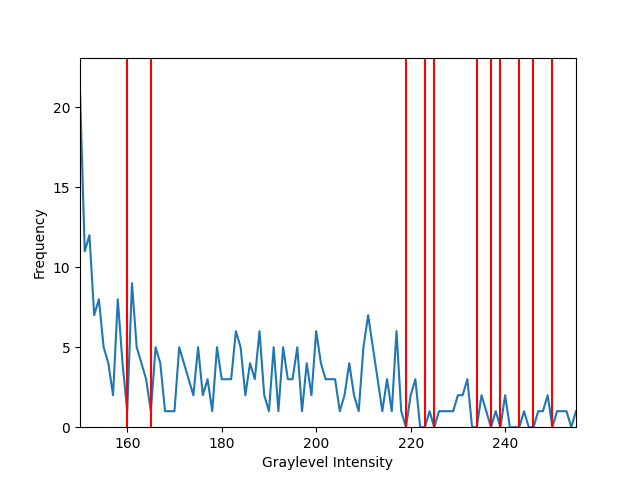}
    \caption{Snippet of the histogram of image \textsc{135069} in the region of grayscale intensities [150,255], showing the thresholds outputted by MET-DP method inside this region.}
    \label{fig:snip-135069}
\end{figure}

For image $\textsc{135069}$, the histogram in \cref{fig:hs_1} is apparent to have two modes. However, MET-DP
produces 13 thresholds. This can be considered as overthresholding since in the plotted histogram we cannot see what
the additional thresholds separate. \Cref{fig:snip-135069} shows a snippet from the histogram for $150 \leq x \leq
255$ and $0 \leq y \leq 23$. We can see here a possible explanation for the extra thresholding done by MET-DP
method. This region contains small fluctuations that can be perceived as 'normal distributions'. This highlights a
sensitivity of MET-DP method to small classes or fluctuations. This can be useful in cases where an image contains objects with different sizes, where Otsu is known to fail \citep{Fan_Lei_2012}.

\begin{figure*}
\plotthresholded{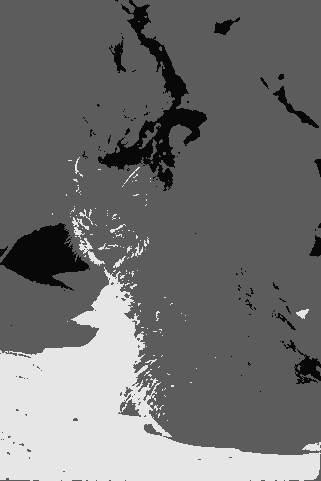}{2} \vspace{0.1cm}
\plotthresholded{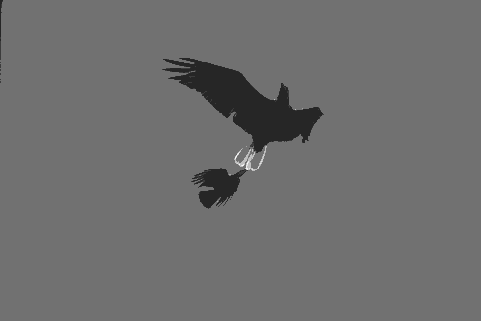}{13} \vspace{0.1cm}
\plotthresholded{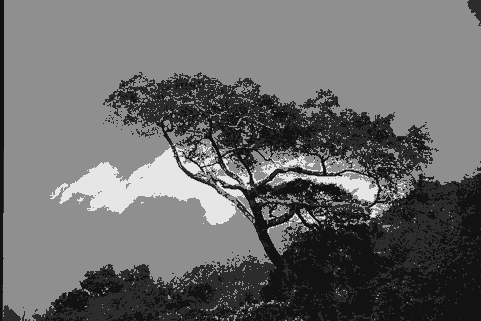}{3} \vspace{0.1cm}
\plotthresholded{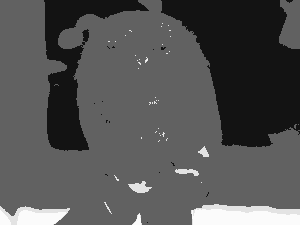}{3} \vspace{0.1cm}
\plotthresholded{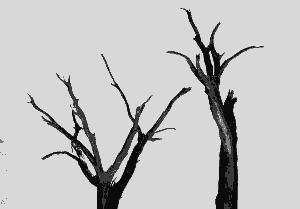}{3}

\caption{Thresholded images obtained by DP\_N\_Otsu, DP\_N\_Kapur, DP\_N\_Kittler and MET-DP methods. From top to bottom, images \texttt{326085} with $n = 2$, \texttt{135069} with $n = 13$, \texttt{147091} with $n = 3$, \texttt{animal-5-bg-020803} with $n = 3$ and \texttt{77200701} with $n = 3$. From left to right, original image, DP\_N\_Otsu, DP\_N\_Kapur, DP\_N\_Kittler and MET-DP.}
\label{fig:th_1}
\end{figure*}

In image $\textsc{147091}$, the histogram in \cref{fig:hs_1} shows 3 or 4 modes. We can see that MET-DP found 3 thresholds. However, these 3 thresholds don't correspond to the 4 modes apparent in the histogram, as we can see that the three bright modes are in the same region (between the second and third threshold), while the two other thresholds separate regions without prominent modes. However, Otsu and Kapur thresholds don't put any thresholds between any of these bright 3 modes. Turning to the original image and its thresholded version in \cref{fig:th_1}. we can see that the three modes correspond to the gray areas of the background/sky. Therefore, merging them in the same thresholded region does not deteriorate significant/important information/features of the image.

The histogram of image $\textsc{animal-5-bg-020803}$, is shown in \cref{fig:hs_1}. We can see that first region resembles a normal distribution, the second region has two normal distributions followed by a linear area, the third region is another linear area while the last region is a normal distribution. The count of 3 thresholds is valid but if the second threshold is close to intensity level 90 and the third threshold is somewhere between 130 and 245. However, none of Otsu, Kapur and Kittler produces such set of thresholds. Hence, the limitation of positioning of these three thresholds as done by MET-DP is shared between these four evaluation functions.

In image $\textsc{77200701}$, the histogram in \cref{fig:hs_1} shows a black peak then a flat area, followed by
one or two mixed normal distributions at region [150,255]. This histogram is best thresholded using a threshold
separating the black peak from other intensities, then an optional threshold (around 150) separating the flat area
from the bright bell curve shape, and an optional threshold around 205 to partially separate the two overlapping normal distributions. We see that Otsu and Kittler are the methods that produce such three thresholds. MET-DP doesn't separate the two overlapping distributions, and splits the flat area instead by an extra threshold around 100. Results of Kapur are similar to those of MET-DP.

\begin{figure*}

\plothistograms{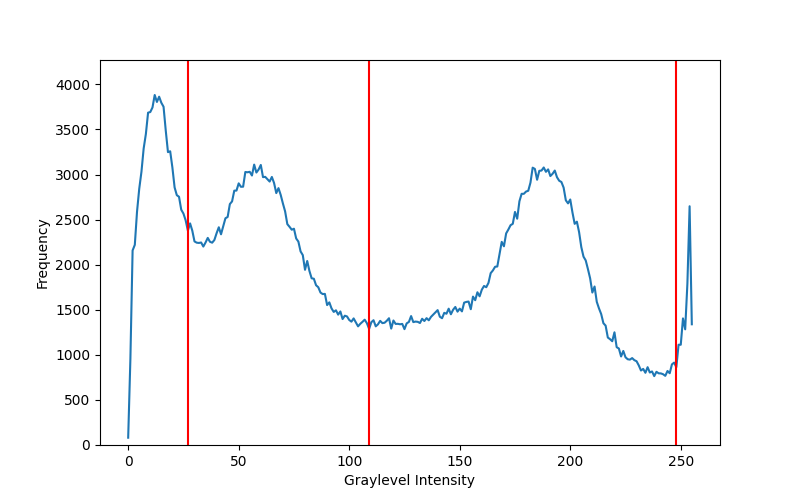}{3}
\plothistograms{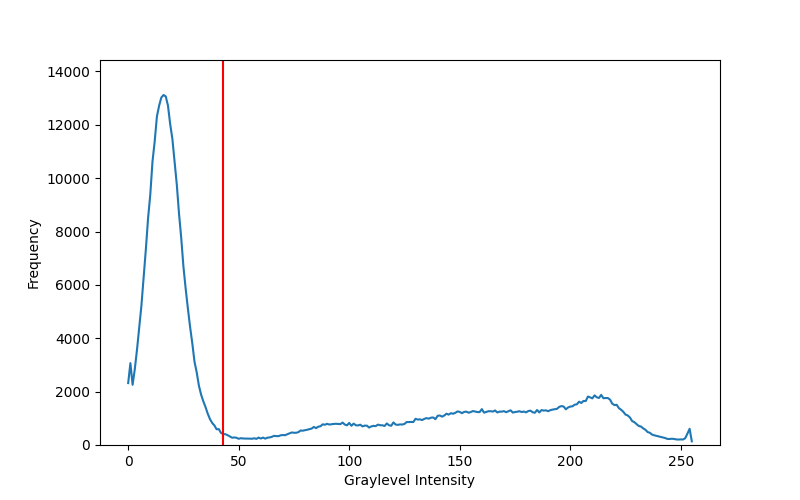}{1}
\plothistograms{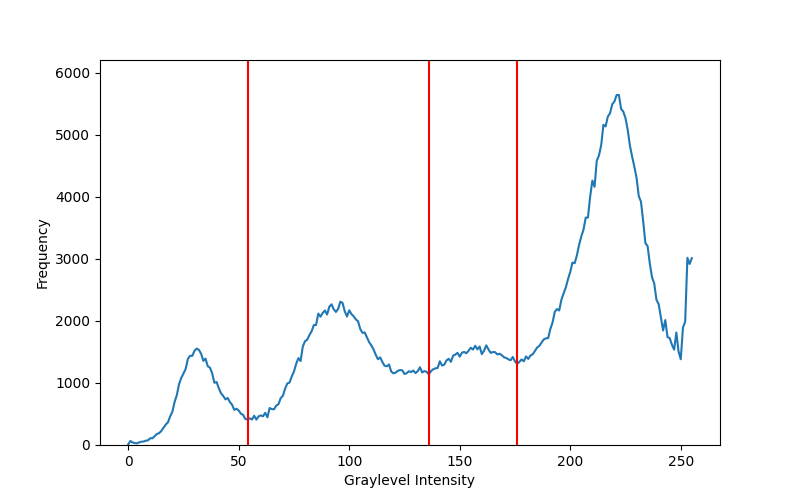}{3}
\plothistograms{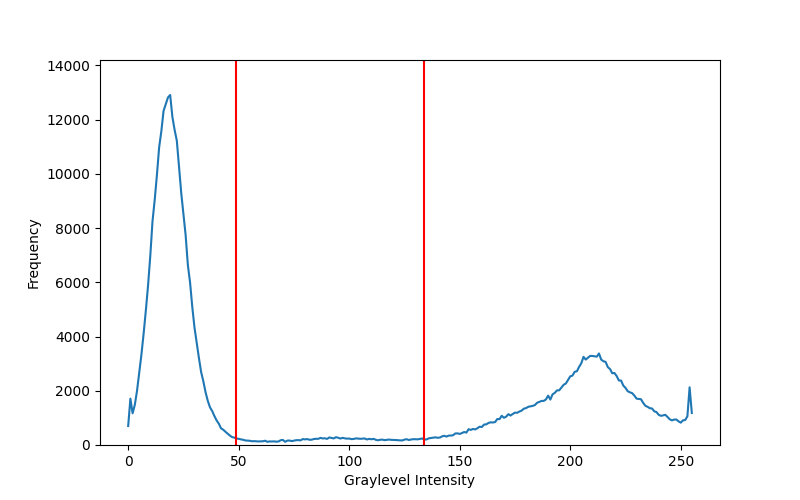}{2}
\plothistograms{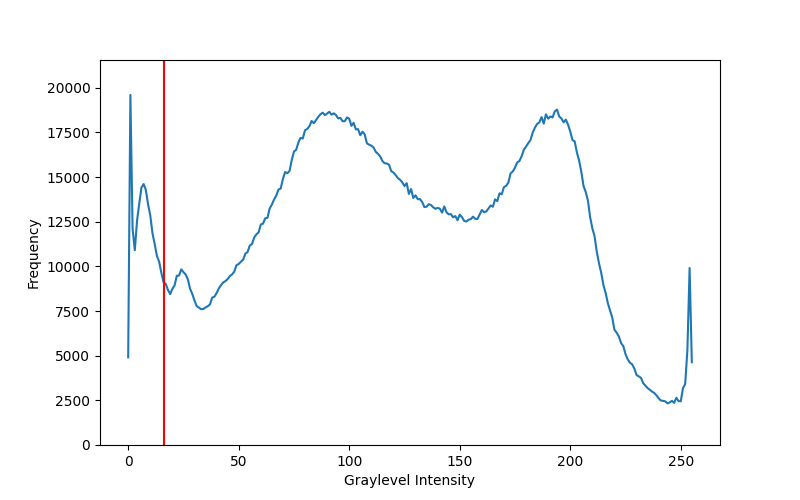}{1}
             \caption{Comparison of DP\_N\_Otsu, DP\_N\_Kapur, DP\_N\_Kittler and MET-DP methods. Each row shows the histogram of a test image with the output thresholds of each methods labeled as vertical red lines. Each column represents the output of one of the four methods. From top to bottom, images \texttt{Tile-1-Part-001} with $n = 3$, \texttt{Tile-3-Part-001} with $n = 1$, \texttt{Tile-3-Part-003} with $n = 3$, \texttt{Tile-3-Part-005} with $n = 2$ and \texttt{Tile-8-Part-008} with $n = 1$. From left to right, DP\_N\_Otsu, DP\_N\_Kapur, DP\_N\_Kittler and MET-DP.}
    \label{fig:hs_2}
\end{figure*}

In image $\textsc{Tile-1-Part-001}$, the histogram in \cref{fig:hs_2} shows 4 clear normal
distributions. Hence, the best number of thresholds is expected to be 3. MET-DP produces 3 thresholds separating
these 4 normal distributions well. Kittler produces a very similar result. However, Otsu thresholds do not separate
the normal distributions as good as the Kittler and MET-DP. Similarly, Kapur merges the first two normal distributions
in one region, and considers the intermediate flat region $[80-148]$ as a separate region. That is why from the perspective of the histogram visual analysis, MET-DP outperforms Otsu and Kapur on this image.

In image $\textsc{Tile-3-Part-001}$, the histogram in \cref{fig:hs_2} shows a dark low-variance normal distribution, followed by a bright high-variance normal distribution. The bright normal distribution can be observed as multiple overlapping normal distributions. However, neither perspective is significantly better than the other. MET-DP determines the best number of thresholds is 1 to separate these two main normal distributions. Its placement by MET-DP is directly after the end of the almost-perfect dark normal distribution. Kittler does the same positioning. However, Otsu and Kapur places the threshold at a brighter intensity level (around 100 for Otsu and 125 for Kapur).

In image $\textsc{Tile-3-Part-003}$, the histogram in \cref{fig:hs_2} consists of 4 normal distributions and a white peak. Hence, we can expect the best number of thresholds to be 4 to separate each of these five regions (four normal distributions and a bright peak). However, the MET-DP method outputs only 3 thresholds by merging the white peak with the closest normal distribution. Kittler shows the same results as well. For Otsu and Kapur, they also separate the four normal distributions. However, the positioning of thresholds is not as accurate as MET-DP. For instance, the first threshold found by Otsu/Kapur lies in the second normal distribution while it could have been placed at the lowest point between these first two normal distributions as apparent in the results of Kittler and MET-DP. The same issue exists on the third threshold (with both Otsu and Kapur).

In image $\textsc{Tile-3-Part-005}$, the histogram in \cref{fig:hs_2} shows two clear normal distributions with a low-weight high-variance normal distribution between them, in addition to a white peak. Hence, the best number of thresholds may be 1 threshold separating the two clear distributions, up to 3 thresholds, separating each of the first normal distribution, the low-weight high-variance one, the last normal distribution and the white peak from each other. MET-DP finds out two thresholds, as it does not separate the white peak from its closest normal distribution. Kapur and Kittler show similar results. However, placement of thresholds determined by Otsu are not as accurate. It includes a part of the low-weight high-variance distribution with its previous one and the other part with its next one. For this last distribution, the first part of it is included with the previous one.

In image $\textsc{Tile-8-Part-008}$, the histogram in \cref{fig:hs_2} shows three small-variance normal distributions followed by two other normal distributions then a white peak. Hence, we can expect a number of thresholds somewhere between 2 and 5. However, MET-DP finds only 1 threshold positioned after the first two small-variance normal distributions. In other words, it considered the first two small-variance distributions as one and considered the third small-variance distribution, the next two distributions and the white peak as only one distribution. Thus, MET-DP is under-thresholding the image $\textsc{Tile-8-Part-008}$.

\begin{figure*}

\plothistograms{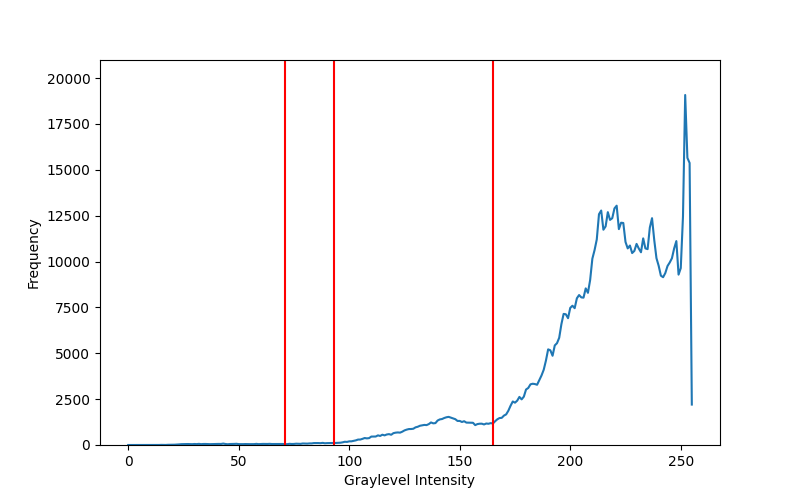}{3}
\plothistograms{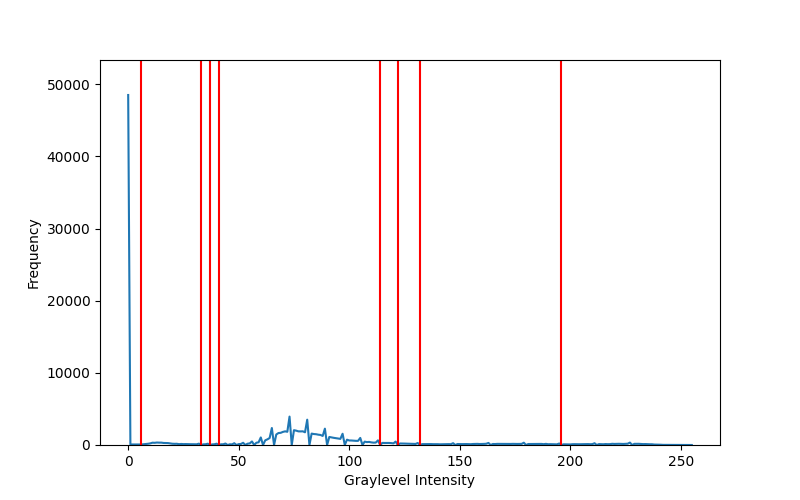}{8}
\plothistograms{007-Flair-60}{25}
\plothistograms{009-Flair-60}{7}
\plothistograms{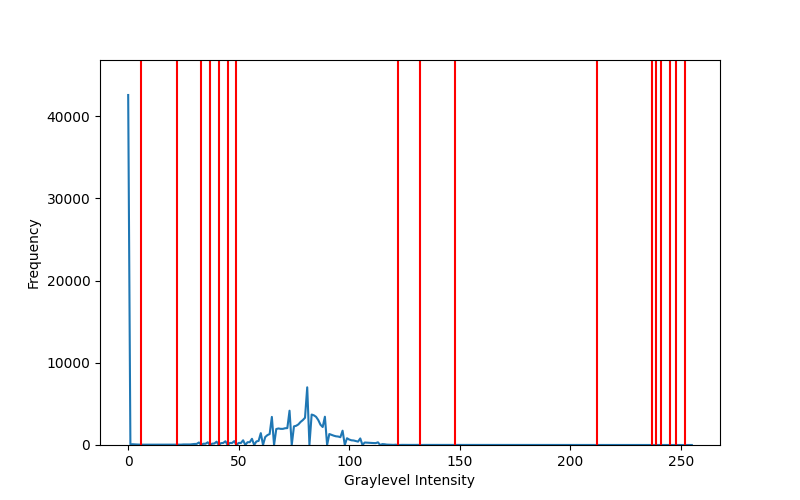}{17}
             \caption{Comparison of DP\_N\_Otsu, DP\_N\_Kapur, DP\_N\_Kittler and MET-DP methods. Each row shows the histogram of a test image with the output thresholds of each methods labeled as vertical red lines. Each column represents the output of one of the four methods. From top to bottom, images \texttt{ISIC-0009880} with $n = 3$, \texttt{006-Flair-60} with $n = 8$, \texttt{007-Flair-60} with $n = 25$, \texttt{009-Flair-60} with $n = 7$ and \texttt{015-t1ce-60} with $n = 17$. From left to right, DP\_N\_Otsu, DP\_N\_Kapur, DP\_N\_Kittler and MET-DP.}
    \label{fig:hs_3}
\end{figure*}

\begin{figure*}
\plotthresholded{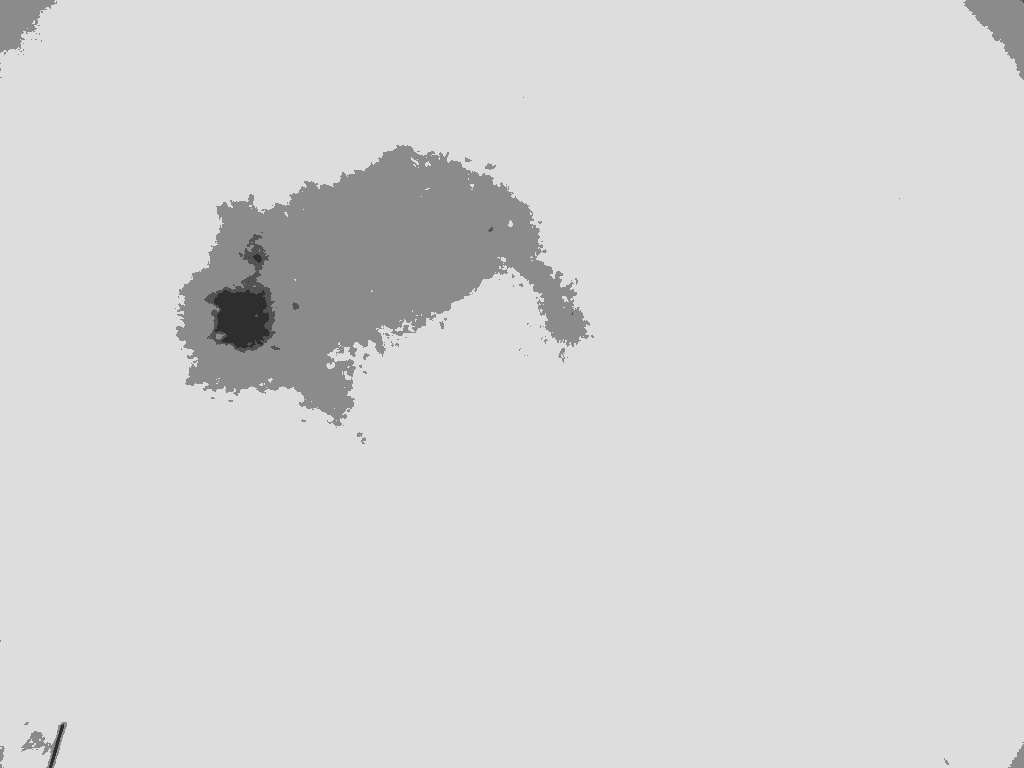}{3} \vspace{0.1cm}
\plotthresholded{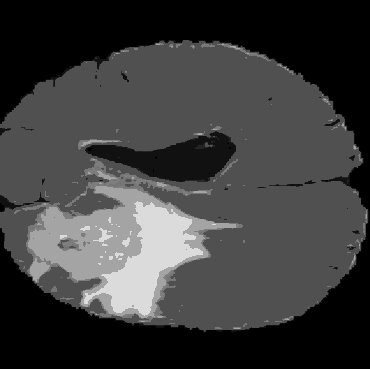}{8} \vspace{0.1cm}
\plotthresholded{007-Flair-60}{25} \vspace{0.1cm}
\plotthresholded{009-Flair-60}{7} \vspace{0.1cm}
\plotthresholded{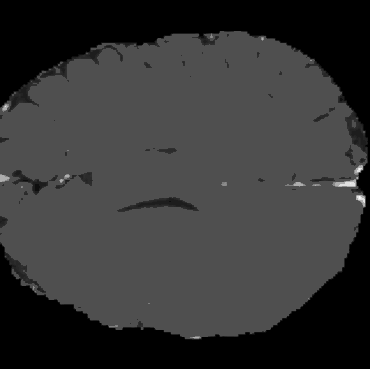}{17}

\caption{Thresholded images obtained by DP\_N\_Otsu, DP\_N\_Kapur, DP\_N\_Kittler and MET-DP methods. From top to bottom, images \texttt{ISIC-0009880} with $n = 3$, \texttt{006-Flair-60} with $n = 8$, \texttt{007-Flair-60} with $n = 25$, \texttt{009-Flair-60} with $n = 7$ and \texttt{015-t1ce-60} with $n = 17$. From left to right, original image, DP\_N\_Otsu, DP\_N\_Kapur, DP\_N\_Kittler and MET-DP.}
\label{fig:th_3}
\end{figure*}

In image $\textsc{ISIC-0009880}$, the histogram in \cref{fig:hs_3} shows overall low frequency for dark pixels while most of the image pixels have bright intensities. Since this image is taken from \ac{isic2016} dataset \citep{isic2016} for skin lesion segmentation, the dark area in the image is the lesion area and is important to distinguish from other areas of the image, as can be seen in \cref{fig:th_3}. MET-DP method figures out three thresholds for this image, resulting in four regions. The first region contains black/very dark pixels, the second region is a small area of dark pixels, the third region is dark gray pixels and the fourth region is the bright/middle gray pixels, which occupies most of the image. The first three regions are mostly part of the lesion area, and the fourth region is mostly non-lesion area. Hence, the choice of the count of three thresholds is suitable here since it maintains important information of the image while avoiding an unnecessarily large number of thresholds.

\begin{figure}
    \centering
    \includegraphics[width=0.9\linewidth]{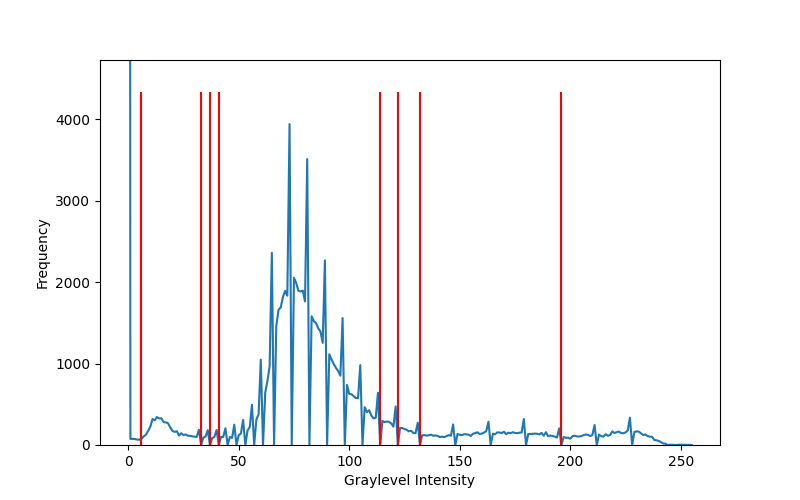}
    \caption{The histogram of image \textsc{006-Flair-60}, the full height of the black peak (at intensity=0) is truncated from the figure so as to clarify other parts of the histogram.}
    \label{fig:snip-006-Flair-60}
\end{figure}

\begin{figure}
    \centering
    \includegraphics[width=0.9\linewidth]{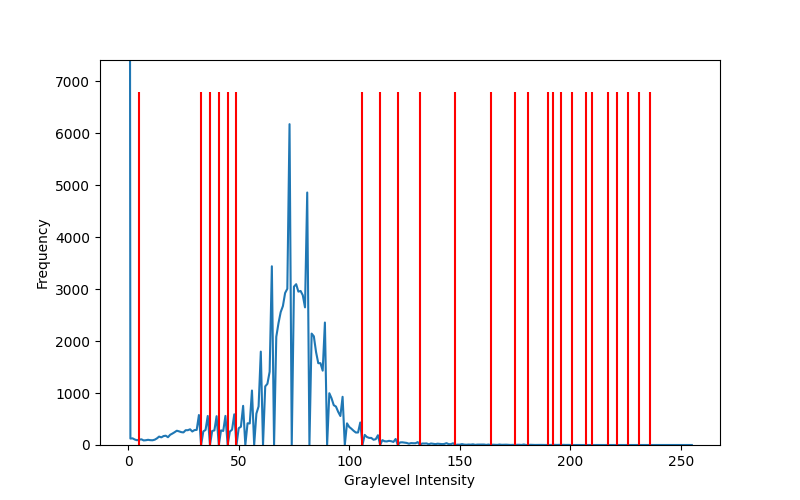}
    \caption{The histogram of image \textsc{007-Flair-60}, the full height of the black peak (at intensity=0) is truncated from the figure so as to clarify other parts of the histogram.}
    \label{fig:snip-007-Flair-60}
\end{figure}
\begin{figure}
    \centering
    \includegraphics[width=0.9\linewidth]{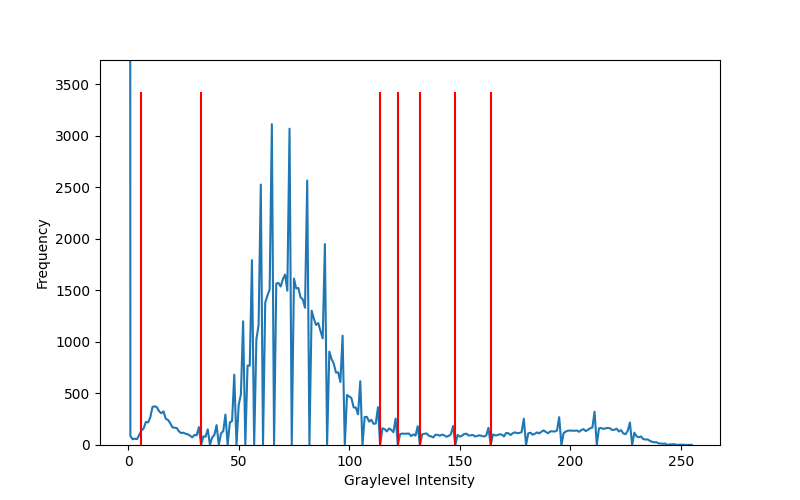}
    \caption{The histogram of image \textsc{009-Flair-60}, the full height of the black peak (at intensity=0) is truncated from the figure so as to clarify other parts of the histogram.}
    \label{fig:snip-009-Flair-60}
\end{figure}

\begin{figure}
    \centering
    \includegraphics[width=0.9\linewidth]{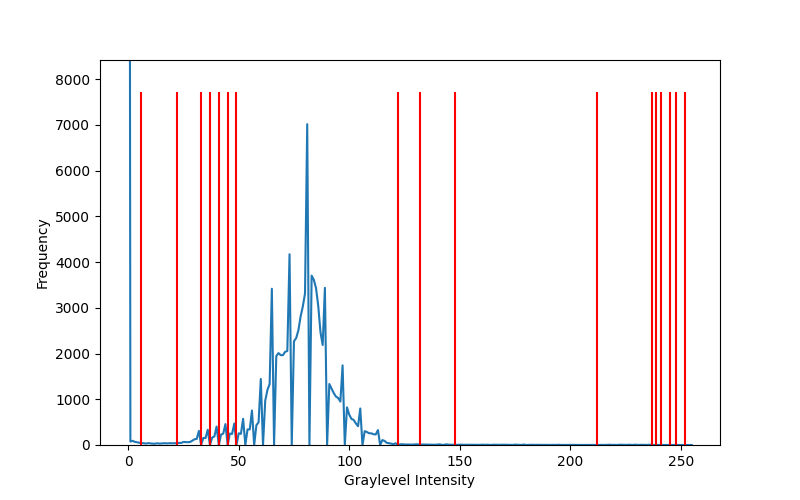}
    \caption{The histogram of image \textsc{015-t1ce-60}, the full height of the black peak (at intensity=0) is truncated from the figure so as to clarify other parts of the histogram.}
    \label{fig:snip-015-t1ce-60}
\end{figure}

\begin{figure*}[h!]
    \plothistogramsn{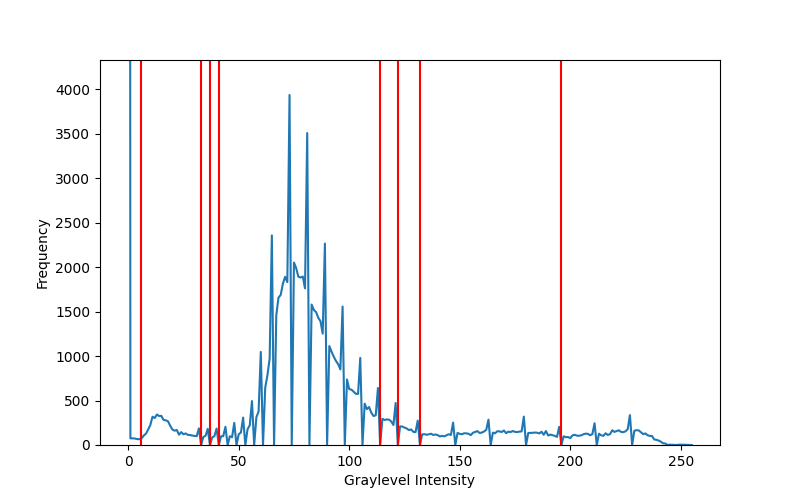}{8}
    \plothistogramsn{007-Flair-60}{25}
    \plothistogramsn{009-Flair-60}{7}
    \plothistogramsn{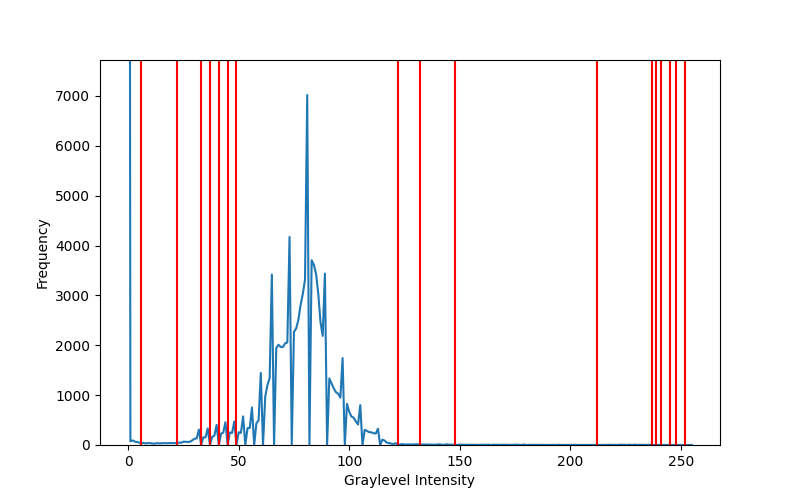}{17}
    \caption{Comparison of methods. The full height of the black peak (at intensity=0) is truncated from the plot so as to clarify other parts of the histogram. Each row shows the histogram of a test image with the output thresholds of each methods labeled as vertical red lines. Each column represents the output of one of the four methods.  From top to bottom, \texttt{006-Flair-60} with $n = 8$, \texttt{007-Flair-60} with $n = 25$, \texttt{009-Flair-60} with $n = 7$ and \texttt{015-t1ce-60} with $n = 17$. From left to right, DP\_N\_Otsu, DP\_N\_Kapur, DP\_N\_Kittler and MET-DP.}
    \label{fig:hsn}
\end{figure*}%

For the images $\{\textsc{006-Flair-60},$ $\textsc{007-Flair-60},$ $\textsc{009-Flair-60},$ $\textsc{015-t1ce-60}\}$, we can see in \cref{fig:hs_3} that all of their histograms have a very large peak at $\text{intensity}=0$. This is justified by the big black background for these images, as can be seen in \cref{fig:th_3}. Therefore and for clarity, we show the same histograms while limiting the plotted frequency to be slightly higher than the maximum frequency of other intensity levels. These histograms are shown in \cref{fig:hsn}


Inspecting the histogram of image \textsc{006-Flair-60} in \cref{fig:hsn}, we can note the following observations. One threshold separated the black pixels from others. Then 3 thresholds separating low-weight fluctuations. The next region is an apparent normal distribution, that has multiple fluctuations inside, but its Gaussian shape is still observable and MET-DP was not biased towards these fluctuations. The next two threshold separate small fluctuations again. The last threshold also separates two regions that can be considered together as one flat region, with multiple fluctuations in it. However, MET-DP was biased towards separating this region into two regions.
Overall, apparent normal distributions in the histogram were distinguished well. However, due to the many fluctuations evident in the histogram, the image was overthresholded to an extent.

In image $\textsc{007-Flair-60}$, the histogram in \cref{fig:hsn}, MET-DP determined the count of thresholds to be 25. The first thresholds separates the black background as one region. For the remaining intensity levels, in the plotted version, we can only see one normal distribution with some fluctuations alongside it. For the intensity levels 125 and higher, the figure does not show their frequencies as they are too small when compared to the normal distribution around the region [50, 100].

In image $\textsc{009-Flair-60}$, the histogram in \cref{fig:hsn} shows the black peak, followed by a
low-weight normal distribution with almost no fluctuations, then a high-weight normal distribution with fluctuations
then a flat area. MET-DP produces 7 thresholds separating the histogram into 8 regions. We can see that the first 3 thresholds separate the histogram into the four regions we described. However, the last four thresholds separate the flat area according to its fluctuations into five smaller regions.

The histogram  for image $\textsc{015-t1ce-60}$ is shown in \cref{fig:hsn}. Here, MET-DP results in 17
thresholds. We can see from the plot of the histogram the black peak and a high-weight normal distribution while the
other regions of the histogram are not clearly distinguishable with respect to the aforementioned normal distribution. Hence,
we cannot reflect deeper in the justification for the 17 found thresholds, similar to results on the image $\textsc{007-Flair-60}$.

\subsubsection{Runtime}

\begin{table*}
    \centering

\begin{tabular}{|c|c|c|c|c|c|}
\hline
Type & Image & Otsu & Kapur & Kittler & MET-DP  \\ \hline
\multirow{5}{*}{Natural} & 326085 & 4.406 & 9.705 & 8.576 & 2.852 \\
& 135069 & 3.962 & 9.402 & 8.454 & 2.605 \\
& 147091 & 4.266 & 9.662 & 8.447 & 2.671 \\
& animal\_5\_bg\_020803 & 4.128 & 10.464 & 8.531 & 2.628 \\
& 7720070 & 4.200 & 9.746 & 8.832 & 2.720 \\
\hline
\multirow{5}{*}{Satellite} & Tile1Part9 & 4.272 & 9.751 & 9.137 & 2.853 \\
& Tile3Part1 & 4.269 & 9.815 & 8.618 & 2.896 \\
& Tile3Part3 & 4.442 & 9.863 & 8.834 & 2.768 \\
& Tile3Part5 & 4.165 & 9.724 & 8.781 & 2.820 \\
& Tile8Part8 & 4.236 & 9.553 & 8.338 & 2.757 \\
\hline
\multirow{5}{*}{Medical} & ISIC\_0009880 & 4.232 & 9.612 & 8.876 & 2.714 \\
& 006\_flair\_60 & 4.076 & 9.745 & 8.658 & 2.587 \\
& 007\_flair\_60 & 4.100 & 9.780 & 8.528 & 2.736 \\
& 009\_flair\_60 & 4.320 & 9.767 & 8.555 & 2.760 \\
& 015\_t1ce\_60 & 4.291 & 9.875 & 8.637 & 2.911 \\
\hline
& Average & 4.205 & 9.783 & 8.601 & 2.738 \\
& Standard Deviation & 0.122 & 0.241 & 0.150 & 0.100 \\
\hline
\end{tabular}
    \caption{CPU time (in seconds) for the 15 test images using DP\_n\_Otsu, DP\_n\_Kapur, DP\_n\_Kittler and MET-DP. The reported CPU time for DP\_N\_Otsu, DP\_n\_Kapur and DP\_n\_Kittler is the total CPU time taken for running each algorithm with all numbers of thresholds between 1 and 15, inclusively. CPU time for MET-DP is for only one run which figures out an appropriate number of thresholds.}
    \label{tab:res-runtime}
\end{table*}

We can see in \cref{tab:res-runtime} the CPU time taken by each of the tested algorithms on each of the used 15 images. As explained in \cref{crit:runtime} in \cref{subsec:eval-crit}, the runtime reported for Otsu, Kapur and Kittler is the total runtime taken for running the conventional dynamic programming algorithm for counts of thresholds from 1 to 15. We can see that the measured runtime values did not change significantly for different tested images. This is evident from the low values of standard deviation of CPU time for each method over the 15 images. We can see that MET-DP had the lowest total CPU runtime among the four methods. Then, Otsu is the second fastest. We can see that Kittler comes between Otsu and Kapur. Since the search method is the same for these three algorithms (conventional dynamic programming which takes the number of thresholds as input), the difference in their runtime is caused by the different computation method of each of them as a mathematical function. While the computation of Otsu function involves computing the zeroth and first-order cumulative moments of the histogram, the computation of Kittler involves computing these and also the second-order cumulative moment of the histogram \citep{otsu_threshold_1979,kittler}. Moreover, two logarithm computations are involved for each Kittler function computation. For Kapur criterion, it involves computing the cumulative first-order logarithmic moment of the histogram \citep{kapur_new_1985}. Therefore, we can see that Kapur is the most complex function out of them to compute while Otsu is the simplest.

\begin{figure*}
        \centering
        \includegraphics[width=0.8\linewidth]{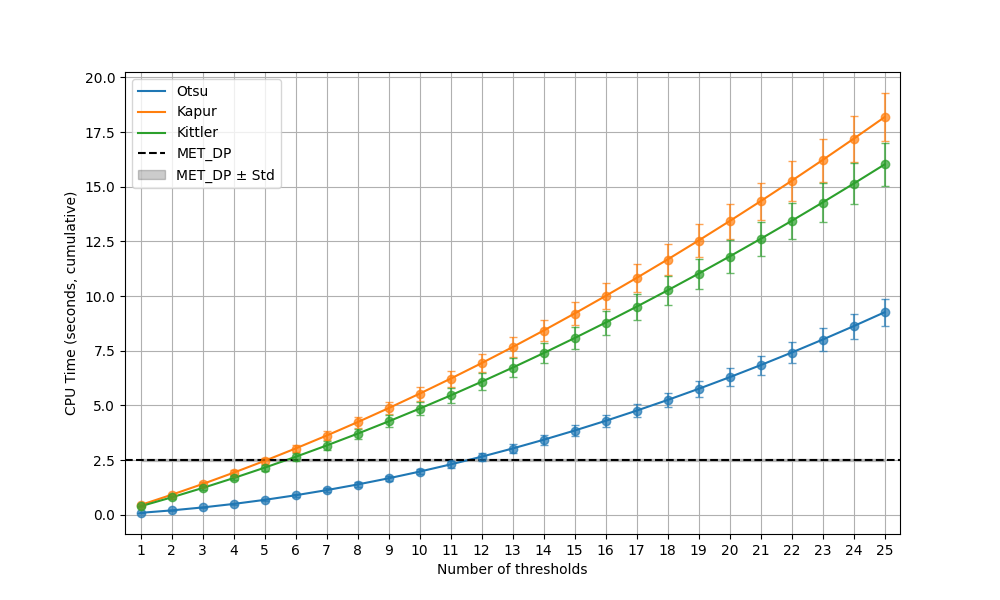}
        \caption{Comparison between the CPU time taken by MET-DP algorithm and conventional dynamic programming algorithm utilizing Otsu, Kapur and Kittler criterion. On y-axis shown is the cumulative CPU time taken for running the conventional dynamic programming technique starting from $n=1$ up to each value of $n$. For each $n$, we show the mean cumulative CPU time and its standard deviation using error bars. For MET-DP, since it does not take $n$ as input, we show only the average CPU time and its standard deviation.}
        \label{fig:cpu-time-cumul}
    \end{figure*}

            For a deeper validation of the runtime comparison, we perform another statistical analysis over a larger number of images. On a dataset of 790 images (500 images from \ac{bsds500}, 200 images from Weizmann database, 72 images from \ac{ssd-uae}, and 18 standard images), we run the conventional dynamic programming algorithm utilizing Otsu, Kapur, Kittler objective functions with values of $n$ ranging from $1$ to $25$ inclusively. We record the CPU time taken by each of them. We also run the MET-DP algorithm on each of these images and record its runtime. Then, we calculate the mean average CPU time and standard deviation of each algorithm over all images given the same number of thresholds. For MET-DP, we record the mean average CPU time and standard deviation over all images. Since the runtime of MET-DP can't be compared to the traditional dynamic programming based methods, we compare its runtime with total runtime of the traditional methods over multiple counts of thresholds. We compute the cumulative CPU time taken by the conventional methods starting at $n=1$ up to each integer value of $n$ in the interval $[1,25]$. \Cref{fig:cpu-time-cumul} shows the findings of this experiment. We see that running the conventional dynamic programming algorithm with each of Kapur and Kittler from $n=1$ till $n=5$ takes less time than MET-DP. Similarly, running it with Otsu up to $n=11$ takes less time than MET-DP. However, starting from $n=6$ for Kapur and Kittler and from $n=12$ for Otsu, the conventional dynamic programming method runtime starts to grow (polynomially) while MET-DP still has the same fixed runtime. This is consistent with the analysis presented in \cref{subsubsec:time-complexity} where it was shown to depend only on $L$, the number of allowed intensity levels. In all the tested images in this article, $L=256$ is fixed. Hence, MET-DP runtime does not change based on the input image content or features. There is still an insignificant variance in runtime values of MET-DP due to other factors of the used machine (memory usage, CPU scheduling and other processes running concurrently).

\begin{figure*}
\plotthresholded{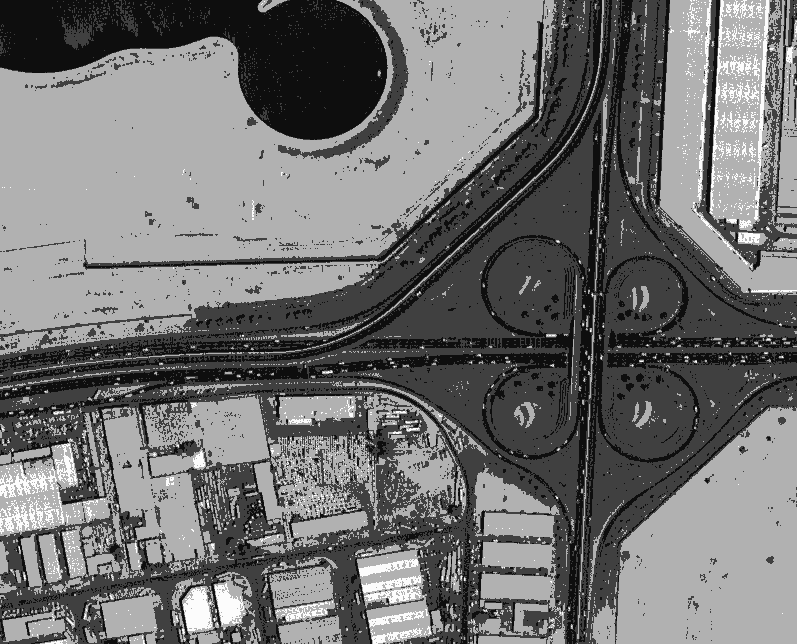}{3} \vspace{0.1cm}
\plotthresholded{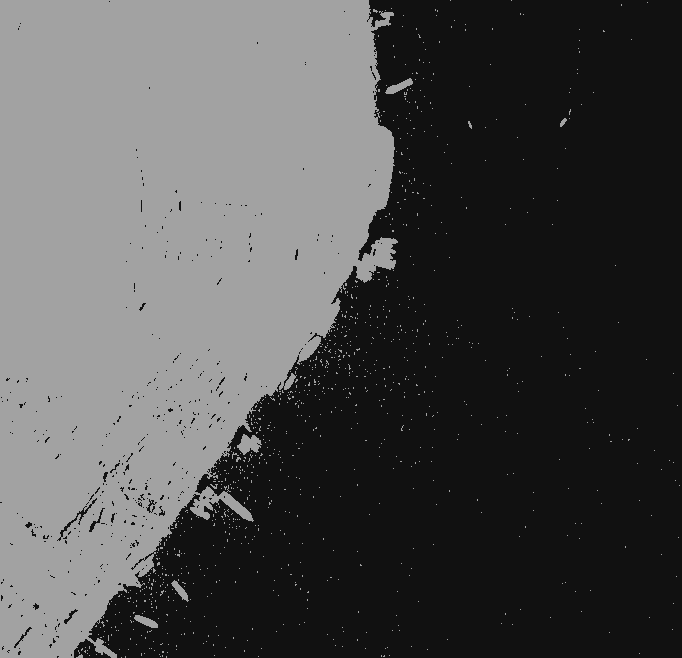}{1} \vspace{0.1cm}
\plotthresholded{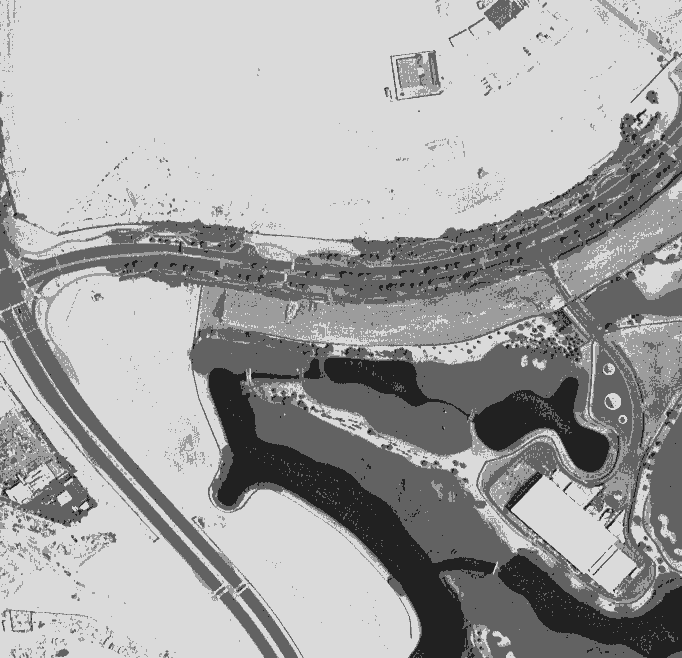}{3} \vspace{0.1cm}
\plotthresholded{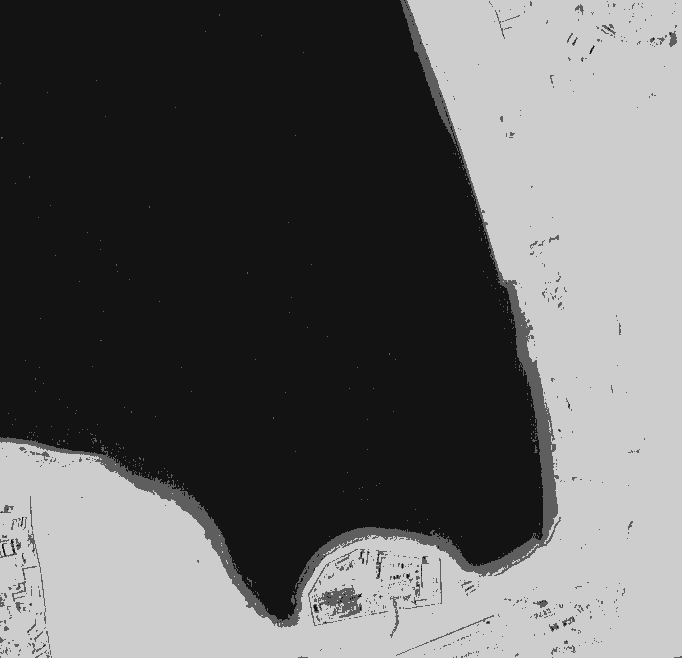}{2} \vspace{0.1cm}
\plotthresholded{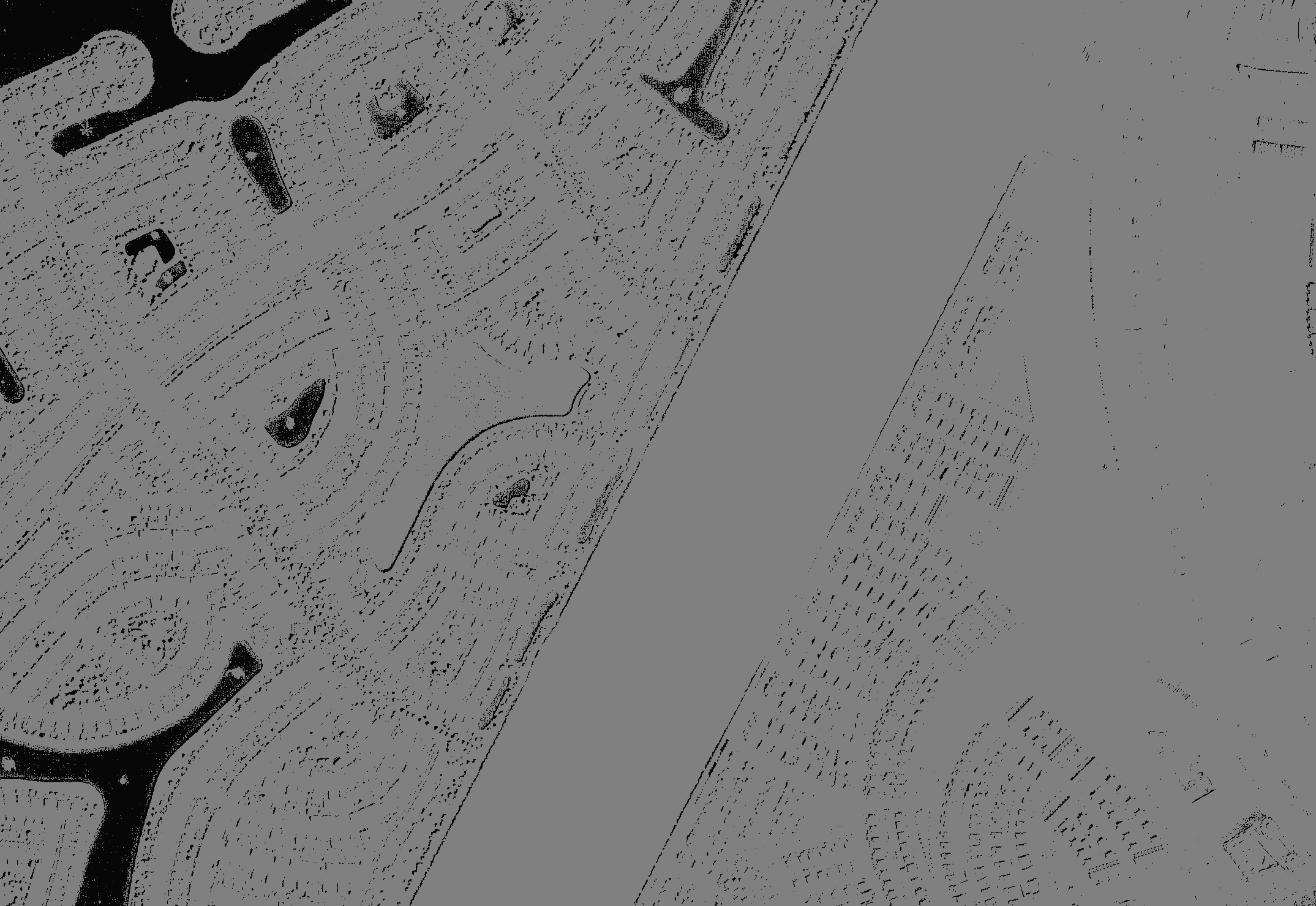}{1}

\caption{Thresholded images obtained by DP\_N\_Otsu, DP\_N\_Kapur, DP\_N\_Kittler and MET-DP methods.
From top to bottom, images \texttt{Tile-1-Part-001} with $n = 3$, \texttt{Tile-3-Part-001} with $n = 1$, \texttt{Tile-3-Part-003} with $n = 3$, \texttt{Tile-3-Part-005} with $n = 2$ and \texttt{Tile-8-Part-008} with $n = 1$. From left to right, original image, DP\_N\_Otsu, DP\_N\_Kapur, DP\_N\_Kittler and MET-DP.}
\label{fig:th_2}
\end{figure*}

\subsubsection{Thresholding Quality}

\begin{table*}[]
    \centering
    \begin{tabular}{|c|c|c|c|c|c|}
    \hline
         Type & Image & Otsu & Kapur & Kittler & MET-DP  \\ \hline
         \multirow{5}{*}{Natural} & 326085 & 0.592 & 0.610 & 0.354 & 0.360 \\
    & 135069 & 0.978 & 0.952 & 0.979 & 0.978 \\
    & 147091 & 0.753 & 0.768 & 0.785 & 0.791 \\
    & animal\_5\_bg\_020803 & 0.764 & 0.763 & 0.549 & 0.547 \\
    & 77200701 & 0.845 & 0.859 & 0.845 & 0.865 \\
    \hline
    \multirow{5}{*}{Satellite} & Tile1Part9 & 0.715 & 0.664 & 0.585 & 0.589 \\
    & Tile3Part1 & 0.468 & 0.464 & 0.423 & 0.423 \\
    & Tile3Part3 & 0.634 & 0.636 & 0.605 & 0.599 \\
    & Tile3Part5 & 0.581 & 0.527 & 0.514 & 0.514 \\
    & Tile8Part8 & 0.523 & 0.524 & 0.225 & 0.231 \\
    \hline
    \multirow{5}{*}{Medical} & ISIC\_0009880 & 0.829 & 0.866 & 0.843 & 0.886 \\
    & 006\_flair\_60 & 0.896 & 0.852 & 0.787 & 0.782 \\
    & 007\_flair\_60 & 0.993 & 0.971 & 0.963 & 0.802 \\
    & 009\_flair\_60 & 0.898 & 0.382 & 0.752 & 0.767 \\
    & 015\_t1ce\_60 & 0.984 & 0.928 & 0.969 & 0.844 \\
    \hline
    \end{tabular}
    \caption{\ac{ssim} values for the 15 test images using DP\_n\_Otsu, DP\_n\_Kapur, DP\_n\_Kittler and MET-DP}
    \label{tab:res-ssim}
\end{table*}

\Cref{tab:res-ssim} shows the \ac{ssim} values obtained when applying the set of thresholds obtained by the MET-DP method on the 15 test images, and the corresponding \ac{ssim} values obtained by the conventional dynamic programming algorithm when used with Otsu, or Kapur, or Kittler using the same number of thresholds derived by the MET-DP method.

For the category of natural images, the MET-DP method obtained superior \ac{ssim} values on the two images $\{\textit{147091, 77200701}\}$ than the three other methods. For the image $\textit{135069}$, the \ac{ssim} value obtained by MET-DP is higher than that of Kapur, same as Otsu and slightly less than Kittler. However, in the remaining two natural images $\{\textit{326085, animal\_5\_bg\_020803} \}$, we see that Otsu and Kapur methods obtain higher \ac{ssim} values than both Kittler and MET-DP.

For the category of satellite images, we can see that at the image $\textit{Tile3Part3}$, MET-DP has the lowest \ac{ssim} values among the four compared methods. In the two images $\{\textit{Tile3Part1, Tile3Part5} \}$, MET-DP and Kittler has the same \ac{ssim} values which are lower than both Otsu and Kapur. For the remaining two images $\{\textit{Tile1Part9, Tile8Part8} \}$, MET-DP has a higher \ac{ssim} value than Kittler, but still lower than Otsu and Kapur.

Turning to the three medical images $\{ISIC\_0009880,$ $006\_flair\_60,$ $009\_flair\_60\}$, we can observe the following. MET-DP produces the highest \ac{ssim} value for the image $\textit{ISIC\_0009880}$. Nonetheless, at the image $\textit{009\_flair\_60}$, it produces the second highest \ac{ssim} value after Otsu. However, at the image $\textit{006\_flair\_60}$, MET-DP produces the lowest \ac{ssim} value among all the four compared methods.

\begin{table*}[]
    \centering
    \begin{tabular}{|c|c|c|c|c|c|}
    \hline
         Type & Image & Otsu & Kapur & Kittler & MET-DP  \\ \hline
         \multirow{5}{*}{Natural} & 326085 & 20.986 & 20.420 & 17.515 & 17.631 \\
& 135069 & 42.301 & 36.795 & 27.109 & 27.079 \\
& 147091 & 24.062 & 24.024 & 19.905 & 20.138 \\
& animal\_5\_bg\_020803 & 24.289 & 23.621 & 17.877 & 17.823 \\
& 77200701 & 26.997 & 23.304 & 26.123 & 23.158 \\
\hline
\multirow{5}{*}{Satellite} & Tile1Part9 & 22.818 & 22.242 & 19.443 & 19.602 \\
& Tile3Part1 & 19.216 & 18.746 & 17.784 & 17.784 \\
& Tile3Part3 & 24.057 & 24.007 & 23.417 & 23.211 \\
& Tile3Part5 & 24.082 & 23.173 & 22.927 & 22.928 \\
& Tile8Part8 & 17.570 & 17.569 & 12.967 & 13.019 \\
\hline
\multirow{5}{*}{Medical} & ISIC\_0009880 & 27.565 & 23.084 & 22.630 & 21.622 \\
    & 006\_flair\_60 & 33.537 & 31.667 & 27.675 & 27.421 \\
    & 007\_flair\_60 & 46.483 & 41.136 & 35.652 & 29.699 \\
    & 009\_flair\_60 & 32.747 & 18.909 & 25.995 & 26.701 \\
    & 015\_t1ce\_60 & 43.840 & 37.429 & 36.300 & 28.264 \\
    \hline
    \end{tabular}
    \caption{\ac{psnr} values for the 15 test images using DP\_n\_Otsu, DP\_n\_Kapur, DP\_n\_Kittler and MET-DP}
    \label{tab:res-psnr}
\end{table*}

\Cref{tab:res-psnr} shows the \ac{psnr} values obtained by the MET-DP method and by the conventional dynamic
programming algorithm with each of Otsu, Kapur, and Kittler objective functions using the number of thresholds derived by the MET-DP method.
We can see that MET-DP method did not achieve the highest \ac{psnr} value on any of the tested images. Moreover,
Otsu always achieved higher \ac{psnr} values than MET-DP on all images. Kapur also achieved higher \ac{psnr} values in all images except the $\textit{009\_flair\_60}$ image where MET-DP excelled over Kapur and Kittler but not Otsu.
Comparing MET-DP with Kittler, we can see that both achieved the same \ac{psnr} value on one image, $\textit{Tile3Part1}$. Other than that, MET-DP gave higher \ac{psnr} values than Kittler on 6 images and lower values on the remaining 6 images.
Hence, we can deduce that MET-DP method does not improve the \ac{psnr} metric values when compared to Otsu or Kapur. However, in some cases, it may achieve better \ac{psnr} values when compared to Kittler.

\section{Discussion}
\label{sec:discussion}

The \textsc{MET-DP} method was extensively evaluated on a diverse dataset of natural, satellite, and medical images. Unlike Otsu, Kapur, and Kittler, which rely on the conventional dynamic programming framework and require the number of thresholds as input, \textsc{MET-DP} employs a different dynamic programming variant that automatically determines the number of thresholds. Its behavior was analyzed through detailed case studies by comparing its threshold placements with the visual structure of histograms and the resulting segmented images.

In images where histogram modes were well-separated, such as \textsc{Tile-1-Part-001} and \textsc{Tile-3-Part-001}, \textsc{MET-DP} successfully identified count and values of thresholds that matched the apparent Gaussian-like distributions, producing accurate and visually meaningful segmentations. However, in some cases, \textsc{MET-DP} showed over-thresholding tendencies, such as in image \textsc{135069} and \textsc{007-Flair-60}, where minor histogram fluctuations were separated into many thresholds. Conversely, instances of under-thresholding were observed, such as in \textsc{Tile-8-Part-008}, where multiple small-variance peaks were merged into broader regions.

In terms of runtime, \textsc{MET-DP} demonstrated a significant advantage by completing a single adaptive search, whereas conventional dynamic programming methods required a number of iterations proportional to the input number of thresholds. For segmentation quality, \textsc{MET-DP} achieved competitive SSIM values in several images, particularly those with complex or multimodal histograms, but its PSNR scores were generally lower than those obtained by Otsu and Kapur. This highlights that \textsc{MET-DP} is especially suited for applications requiring automatic threshold selection and computational efficiency, as well as scenarios where separating fine histogram fluctuations or subtle intensity variations is important. However, for tasks prioritizing high PSNR or coarse, noise-tolerant segmentations, traditional objective-based methods may still perform better.

\section{Conclusion}
\label{sec:conclusion}

This paper presented an extensive explanation and experimental evaluation of the \textsc{MET-DP} method for automatic multi-level image thresholding. While \textsc{MET-DP} is not proposed in this work, its algorithmic principles, computational properties, and segmentation performance were analyzed in depth and compared with \textsc{Otsu}, \textsc{Kapur}, and \textsc{Kittler}, which employ the conventional dynamic programming algorithm. In contrast, \textsc{MET-DP} utilizes a different dynamic programming variant, enabling automatic estimation of the number of thresholds without requiring prior specification.

A statistical study was conducted to investigate the effect of modifying the objective functions of Otsu, Kapur, and Kittler in the same fashion used to derive the \textsc{MET-DP} formulation from the original Kittler criterion. Results indicate that the modified Kittler-based \textsc{MET-DP} criterion exhibits a significantly improved response to different numbers of thresholds, making it suitable for automatically selecting the optimal threshold count by comparing function values. Conversely, similar modifications applied to \textsc{Otsu} and \textsc{Kapur} did not produce meaningful changes, as their objective values remained biased toward higher numbers of thresholds, limiting their usefulness for automated selection.

The paper also provided a complexity analysis of \textsc{MET-DP}. The algorithm has a time complexity of $O(L^3)$ and two implementation variants with an asymptotic space complexity of $O(L^2)$. The optimized version achieves reduced memory usage by replacing a 2D ``indices'' array with a 1D representation while maintaining correctness, resulting in better practical space efficiency.

To illustrate \textsc{MET-DP}'s behavior, an experiment on a sample image  demonstrated how transforming the Kittler criterion into the \textsc{MET-DP} formulation reduced a severe over-thresholding case to a much more meaningful segmentation. A synthetic example was also presented to show, step by step, how the algorithm computes function values, compares candidates, and determines thresholds, further demonstrating the effectiveness of the \textsc{MET-DP} modification.

Finally, experimental results on natural, satellite, and medical images demonstrate that \textsc{MET-DP} consistently achieves the lowest computational runtime, benefiting from its efficient threshold search strategy. In terms of segmentation quality, \textsc{MET-DP} achieves high SSIM on several images, especially where histograms exhibit complex multimodal patterns. However, the algorithm demonstrates sensitivity to small histogram fluctuations, leading to over-thresholding in some natural and medical images and under-thresholding in certain satellite images. While \textsc{MET-DP} performs well perceptually, it typically produces lower PSNR values than Otsu and Kapur, indicating slightly reduced pixel-level fidelity.

In summary, \textsc{MET-DP} provides a computationally efficient, adaptive, and robust thresholding framework capable of automatically estimating the number of thresholds and handling diverse image types. Future work will explore the applicability of the proposed approach on more diverse image modalities. Additionally, applicability of the proposed dynamic programming scheme with other objective functions not explored in this article is to be explored.

%


\bibliographystyle{elsarticle-harv}
\bibliography{references}


\end{document}